\documentclass[10pt,twocolumn,letterpaper]{article}
\newif\ifreviewversion
\reviewversionfalse
\ifreviewversion
  \usepackage[review]{cvpr}
\else
  \usepackage[pagenumbers]{cvpr}
\fi
\def\paperID{0001}
\usepackage{microtype}
\usepackage{float}
\usepackage{array}
\usepackage{colortbl}
\usepackage{tabularx}
\newcommand{\methodname}{Direct Forcing}
\newcommand{\method}{\methodname\xspace}
\newcommand{\sg}{\operatorname{sg}}
\newcommand{\mmd}{\operatorname{MMD}}
\newcommand{\clip}{\operatorname{clip}}
\newcommand{\vm}{VideoMAE v2\xspace}

\definecolor{tablelavender}{RGB}{234,228,245}
\definecolor{grouplight}{RGB}{247,247,249}
\definecolor{placeholderfill}{RGB}{248,247,251}

\makeatletter
\iftoggle{cvprfinal}{}{%
  \@ifpackagelater{lineno}{2026/02/16}{%
    \@ifpackagelater{lineno}{2026/02/17}{}{%
      \IfFormatAtLeastTF{2025-06-01}{%
        \RemoveFromHook{build/column/before}[lineno]
        \AddToHook{build/column/after}[directforcing-lineno]{%
          \setbox\@outputbox\vbox{%
            \boxmaxdepth\@maxdepth
            \protected@write\@auxout{}{%
              \string\def\string\@LN@column{\if@firstcolumn1\else2\fi}%
            }%
            \box\@outputbox
          }%
        }%
      }{}%
    }%
  }{}%
}
\makeatother

\definecolor{cvprblue}{rgb}{0.21,0.49,0.74}
\ifreviewversion
  \usepackage[pagebackref,breaklinks,colorlinks,allcolors=cvprblue]{hyperref}
\else
  \usepackage[breaklinks,colorlinks,allcolors=cvprblue]{hyperref}
\fi
\hypersetup{pdftitle={\methodname: Single-Stage Training with Representation-Space Supervision for Few-Step Autoregressive Video Generation},pdfauthor={}}
\def\confName{CVPR}
\def\confYear{2027}
\title{\methodname: Single-Stage Training with Representation-Space Supervision for Few-Step Autoregressive Video Generation}
\makeatletter
\iftoggle{cvprfinal}{
  \patchcmd{\@maketitle}{\vspace*{24pt}}{\vspace*{8pt}}{}{
    \PackageError{directforcing}{Cannot adjust title-to-author spacing}{}
  }
  \patchcmd{\@maketitle}{\vspace*{12pt}}{\vspace*{8pt}}{}{
    \PackageError{directforcing}{Cannot adjust author-block bottom spacing}{}
  }
}{}
\makeatother
\author{
Haojun Xu$^{1,2}$ \enspace
Jie Huang$^{2,*,\ddagger}$ \enspace
Chuyang Zhao$^{2,\ddagger}$ \enspace
Xin Lu$^{2}$ \enspace
Mingchen Zhong$^{2}$\\[2pt]
Zihao Fan$^{2}$ \enspace
Haoyang Huang$^{2}$ \enspace
Linjiang Huang$^{1,\dagger}$ \enspace
Si Liu$^{1}$\\[2pt]
$^{1}$Beihang University \quad
$^{2}$JD Future Academy\\[2pt]
$^{\dagger}$Corresponding author. \quad
$^{\ddagger}$Project leader. \quad
$^{*}$Co-first author.
}
\begin{document}
\maketitle
\begin{abstract}
Few-step autoregressive video generation enables low-latency streaming, 
yet prevailing training pipelines separate causal adaptation from distribution matching. 
Recent representation-based text-to-video methods retain this separation by starting with an ODE initialization stage, even when the target distribution is defined directly by reference videos.
We introduce \method, the first single-stage framework that uses reference videos both as paired examples of how a history continues and as a target population for self-generated rollouts.
Specifically, \method is trained with two complementary signals: (1) Causal Continuation Alignment learns paired spatiotemporal correspondence under reference histories; and (2) Rollout Distribution Matching employs maximum mean discrepancy to align the representation distributions of self-generated and reference videos.
To coordinate these signals, we introduce Dynamic Gradient Balancing, which adapts their relative gradient contributions during training.
Experiments with Wan2.1-1.3B show that \method learns a four-step autoregressive generator in 250 updates without a separate ODE initialization stage or online diffusion-score supervision, achieving 84.63 VBench Total and 86.01 Quality. Code at \href{https://github.com/xuhaojun1/Direct-Forcing}{Direct Forcing}.
\end{abstract}

\begin{figure}[t]
\centering
\includegraphics[width=\columnwidth]{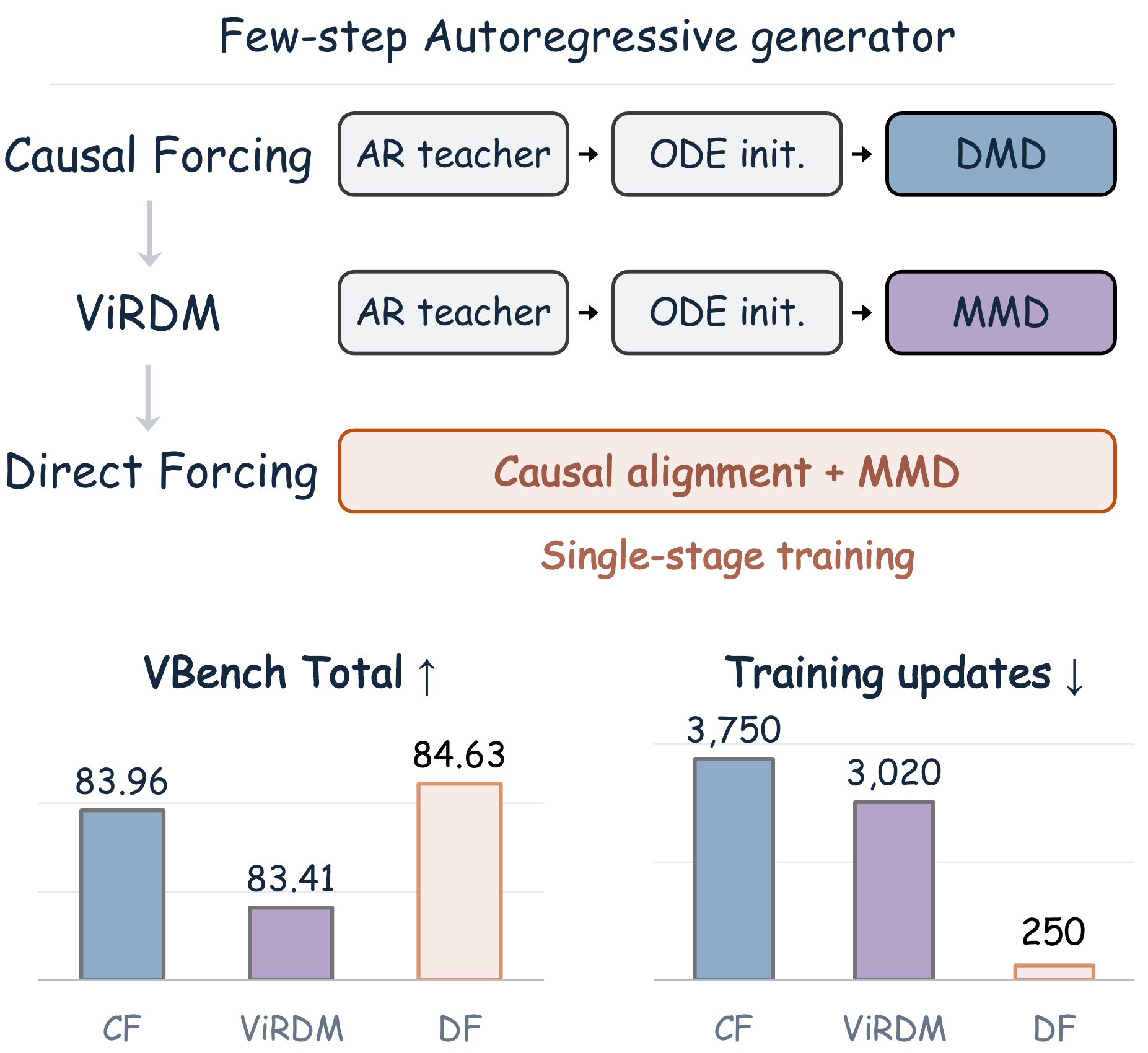}%
\caption{\textbf{Single-stage training for few-step AR generation.} \method jointly learns causal alignment and distribution matching, achieving 84.63 VBench Total in 250 updates. Baseline counts include inherited initialization (Table~\ref{tab:pipeline}).}
\label{fig:teaser}
\end{figure}
\section{Introduction}
\label{sec:intro}

Autoregressive (AR) video diffusion generates temporal chunks conditioned on preceding content, supporting streaming and interactive applications~\cite{causvid,selfforcing,longlive}. However, the computational cost of multi-step diffusion sampling limits real-time generation. To reduce latency, recent work distills pretrained bidirectional video diffusion models into few-step AR students. This adaptation typically follows a two-stage pipeline: initializing AR students using ordinary differential equation (ODE) trajectories to learn few-step causal continuation and narrow the initial student--teacher distribution gap, followed by distribution matching distillation (DMD)~\cite{dmd,dmd2} to improve generation quality. For example, Self Forcing~\cite{selfforcing} and Causal Forcing~\cite{causalforcing} initialize their students using ODE trajectories from bidirectional and causal teachers, respectively.

Instead of using DMD, recent methods compare generated and reference videos in frozen representation spaces to improve few-step generation quality, measuring their distributional discrepancy with maximum mean discrepancy (MMD). Starting from ODE-initialized checkpoints, ViRDM~\cite{virdm} and Elastic Forcing~\cite{elastic} adopt this sample-based objective without requiring an online diffusion teacher or a learned score model for generated samples. However, these pipelines retain a separate AR student initialization stage before distribution matching. Since MMD defines its supervision directly from generated and reference samples, prior alignment with a diffusion teacher is not a prerequisite of the objective. We therefore ask: \emph{Can an AR student learn causal continuation and match the reference-video distribution jointly, without a separate initialization stage?}

\begin{figure}[t]
\centering
\includegraphics[width=\columnwidth]{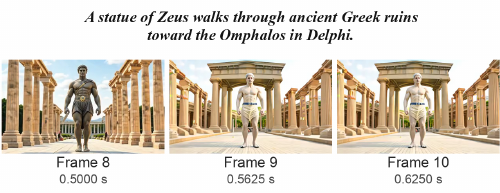}
\caption{\textbf{Temporal discontinuity under MMD-only training.} Consecutive frames 8--10 show an abrupt change at the first chunk boundary.}
\label{fig:mmd-failure}
\end{figure}

\begin{figure*}[t]
\centering
\includegraphics[width=\textwidth]{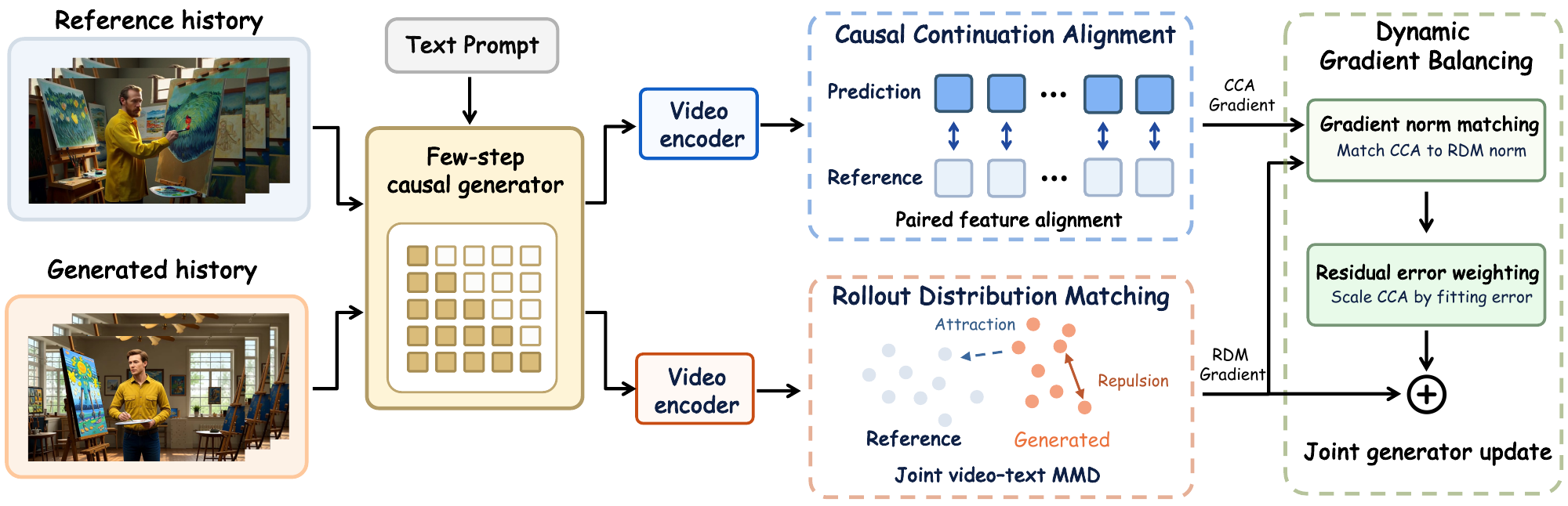}
\caption{\textbf{\method overview.} A shared few-step causal generator learns from reference histories with Causal Continuation Alignment (CCA; Sec.~\ref{sec:tf}) and self-generated histories with Rollout Distribution Matching (RDM; Sec.~\ref{sec:mmd}). Frozen encoders provide paired cosine alignment and video--text MMD, respectively. Dynamic Gradient Balancing (Sec.~\ref{sec:controller}) first matches gradient norms, then weights the CCA gradient by its current fitting residual before combining both gradients to update the shared generator.}
\label{fig:method}
\end{figure*}

We take a new perspective on reference video supervision: each video supplies paired examples of how a history continues, while the collection defines a target population for self-generated rollouts. MMD only uses the population-level signal but does not explicitly supervise the paired history--continuation correspondence. We conduct direct MMD-only training and observe temporal discontinuities at chunk boundaries (Figure~\ref{fig:mmd-failure}), which motivates explicit continuation supervision alongside distribution matching. We therefore introduce \method (Figure~\ref{fig:method}), the first single-stage framework that adapts a pretrained bidirectional video diffusion model into a few-step autoregressive generator directly by learning from reference videos.

Specifically, to learn from paired examples, \emph{Causal Continuation Alignment (CCA)} aligns predicted and reference spatiotemporal features under reference histories. Because these histories do not contain accumulated prediction errors, we also train on complete rollouts generated under the model's own histories. \emph{Rollout Distribution Matching (RDM)} matches the joint video--text representation distribution of these rollouts to that of the references through MMD. Both objectives update a shared generator from the outset (Figure~\ref{fig:method}). We select their frozen representation spaces independently, since paired correspondence and population matching need not favor the same features.

Joint training usually coordinates the strength of the above two signals. Dense paired supervision dominates RDM because of its larger gradient scale. Correcting this imbalance alone does not determine how much causal continuation supervision in CCA to retain as fitting improves. We therefore introduce \emph{Dynamic Gradient Balancing}, which separates gradient-scale calibration from residual-based weighting while preserving a lower bound on the causal continuation contribution.

Starting from pretrained bidirectional Wan2.1-1.3B, \method learns a four-step causal generator in 250 joint updates, achieving 84.63 VBench Total and 86.01 Quality without a separate initialization stage. Table~\ref{tab:pipeline} compares this joint training route with staged recipes, including inherited initialization updates.

\begin{table*}[t]
\centering
\small
\setlength{\tabcolsep}{5pt}
\renewcommand{\arraystretch}{1.13}
\caption{\textbf{Training paths from bidirectional Wan2.1-1.3B to four-step AR generation.} \emph{BID} denotes bidirectional. Numeric entries report optimization updates for causal-teacher training, ODE initialization, and distribution matching; Total includes inherited stages. Our 250 updates jointly optimize Causal Continuation Alignment (CCA) and Rollout Distribution Matching (RDM). Backbone pretraining and offline data/feature preparation are excluded.}
\label{tab:pipeline}
\begin{tabularx}{\textwidth}{>{\raggedright\arraybackslash}p{1.1in}>{\raggedright\arraybackslash}p{1.7in}
>{\centering\arraybackslash}p{1.1in}
>{\hsize=.75\hsize\linewidth=\hsize\centering\arraybackslash}X
>{\hsize=1.75\hsize\linewidth=\hsize\centering\arraybackslash}X
>{\hsize=.5\hsize\linewidth=\hsize\centering\arraybackslash}X}
\toprule
\rowcolor{tablelavender}
\multicolumn{1}{c}{\textbf{Method}} & \multicolumn{1}{c}{\textbf{Training path}} & \textbf{AR teacher training} & \textbf{ODE init.} & \textbf{Distribution matching} & \textbf{Total} \\
\midrule
Self Forcing~\cite{selfforcing,causalforcing} & ODE init. (BID teacher) $\rightarrow$ DMD & -- & 3,000 & 750 & 3,750 \\
Causal Forcing~\cite{causalforcing} & ODE init. (AR teacher) $\rightarrow$ DMD & 2,000 & 1,000 & 750 & 3,750 \\
ViRDM~\cite{virdm} & ODE init. (AR teacher) $\rightarrow$ MMD & 2,000 & 1,000 & 20 & 3,020 \\
Elastic Forcing~\cite{elastic} & ODE init. (BID teacher) $\rightarrow$ MMD & -- & 3,000 & 150 & 3,150 \\
\midrule
\rowcolor{black!4}
\textbf{\method (ours)} & \textbf{Joint CCA + RDM} & \textbf{--} & \textbf{--} & \textbf{250} & \textbf{250} \\
\bottomrule
\end{tabularx}
\end{table*}

Our contributions are:
\begin{itemize}[leftmargin=*,itemsep=2pt,topsep=3pt]
\item \textbf{A single-stage framework for few-step causal video generation.}
\method jointly learns causal continuation and rollout distribution matching from reference videos, adapting a pretrained bidirectional model into a four-step causal generator without a separate initialization stage.

\item \textbf{Dynamic gradient balancing for joint optimization.}
We separate gradient-scale calibration from fitting-dependent weighting to coordinate continuation alignment and rollout distribution matching. Ablations show benefits from both gradient norm matching and residual error weighting.

\item \textbf{Distinct representation requirements for the two objectives.}
Our ablations support selecting representation spaces independently: for VideoMAE v2 and V-JEPA 2.1, deeper features favor paired continuation alignment, whereas intermediate features improve generation quality and Dynamic Degree in rollout distribution matching.
\end{itemize}

\section{Related Work}
\label{sec:related}
\subsection{Few-Step Autoregressive Video Generation}
VideoGPT~\cite{videogpt} and VideoPoet~\cite{videopoet} generate discrete visual tokens, while NOVA~\cite{nova} uses continuous autoregressive generation. Streaming extensions improve cache reuse, noise scheduling, and history conditioning~\cite{ca2vdm,magi,diffusionforcing,skyreelsv2,historyguided,framepack}. For diffusion-based generators, reducing denoising steps further requires dedicated training recipes. CausVid~\cite{causvid} combines ODE initialization with DMD for few-step causal generation. Self Forcing~\cite{selfforcing} and Causal Forcing~\cite{causalforcing} use bidirectional and causal ODE initialization, respectively, followed by DMD on student rollouts. Matrix-Game 3.5~\cite{qian2026matrix} uses teacher-forced perceptual flow matching for few-step causal initialization, then improves self-generated rollouts with DMD. \method instead jointly learns causal continuation and rollout distribution matching in a single stage.

\subsection{Representation Distribution Matching}
Early sample-based generators optimize MMD between generated and reference populations~\cite{gmmn,mmdnets}, using differences between kernel mean embeddings~\cite{gretton2012}. Recent methods compare learned representations through attraction--repulsion fields~\cite{drifting}, feature moments~\cite{rfl,amfd,advfd}, or Sinkhorn-divergence gradient flows~\cite{wflow}. RDM~\cite{rdm} applies MMD under frozen pretrained encoders, allowing reference summaries to be computed offline. ViRDM~\cite{virdm} and Elastic Forcing~\cite{elastic} extend this approach to causal rollouts with memory-efficient differentiation and reference estimation. Despite learning from reference videos, their standard text-to-video recipes retain ODE-initialized checkpoints.

\section{Method}
\label{sec:method}
\method jointly adapts a pretrained bidirectional model for few-step autoregressive generation in a single stage (Figure~\ref{fig:method}). \emph{Causal Continuation Alignment} (CCA; Sec.~\ref{sec:tf}) supervises predictions under reference histories through paired spatiotemporal feature alignment. \emph{Rollout Distribution Matching} (RDM; Sec.~\ref{sec:mmd}) uses MMD to match feature distributions of self-generated rollouts and reference videos. \emph{Dynamic Gradient Balancing} (Sec.~\ref{sec:controller}) controls both objectives' relative gradient contributions to the shared generator.

\subsection{Problem Setup}
\label{sec:setup}
Let $\mathcal D=\{(x_i,c_i)\}_{i=1}^N$ be a dataset of reference video--prompt pairs. Each video is represented in latent space as $x=(x^1,\ldots,x^J)$, where $x^j$ is the $j$-th temporal chunk. The autoregressive distribution factorizes as
\begin{equation}
p_\theta(\hat x\mid c)=\prod_{j=1}^{J}p_\theta(\hat x^j\mid\hat x^{<j},c).
\label{eq:ar}
\end{equation}
Only generator parameters $\theta$ are optimized; feature encoders remain frozen. We use $\sg(\cdot)$ to denote stop-gradient.

\subsection{Causal Continuation Alignment}
\label{sec:tf}
We predict each chunk from its noisy input and clean reference history, then align the resulting video with its paired reference in a spatiotemporal feature space. Given a noisy chunk $x_\sigma$, history condition $h$, and prompt $c$, we define the clean-endpoint predictor as
\[
F_\theta(x_\sigma,\sigma;h,c)
=x_\sigma-\sigma u_\theta(x_\sigma,\sigma;h,c),
\]
where $u_\theta$ is the network's predicted velocity.

For each chunk $j$, we independently sample $\sigma^j$ uniformly from the four denoising levels $\{\sigma_1,\ldots,\sigma_4\}$ and use the same level for all frames within that chunk:
\begin{equation}
x^j_{\sigma^j}=(1-\sigma^j)x^j+\sigma^j\epsilon^j,
\qquad \epsilon^j\sim\mathcal N(0,I).
\label{eq:noise}
\end{equation}
Let $\boldsymbol\sigma=(\sigma^1,\ldots,\sigma^J)$ and $x_{\boldsymbol\sigma}=(x^1_{\sigma^1},\ldots,x^J_{\sigma^J})$. To predict all chunks in one forward pass, we concatenate the clean reference stream $x$ and the noisy stream $x_{\boldsymbol\sigma}$. We apply $F_\theta$ under attention mask $\mathcal M$ and retain only the noisy-stream predictions. The predicted video $\tilde x=(\tilde x^1,\ldots,\tilde x^J)$ is thus
\begin{equation}
\tilde x=F_\theta(x_{\boldsymbol\sigma},\boldsymbol\sigma;x,c).
\label{eq:tf-predict}
\end{equation}
With rows indexing query chunks and columns indexing key chunks, both ordered as [clean; noisy], we use the block mask below. Here, $A,B\in\{0,1\}^{J\times J}$, $I_J$ is the $J\times J$ identity matrix, and $\mathbf 1[\cdot]$ denotes the indicator function:
\begin{equation}
\mathcal M=\begin{bmatrix}A&0\\B&I_J\end{bmatrix},\qquad
\begin{aligned}
A_{jk}&=\mathbf 1[k\le j],\\
B_{jk}&=\mathbf 1[k<j].
\end{aligned}
\label{eq:cca-mask}
\end{equation}
Each clean chunk attends to itself and preceding clean chunks. Each noisy chunk attends only to preceding clean chunks and its own noisy tokens.

We align $\tilde x$ with $x$ using a frozen spatiotemporal video encoder. Let $P$ be the number of aligned token positions, and let $\bar\Phi(x,c)_p$ denote the unit-normalized feature at position $p\in\{1,\ldots,P\}$. The predicted and reference features are $f_p=\bar\Phi(\tilde x,c)_p$ and $f_p^*=\sg(\bar\Phi(x,c)_p)$, respectively. The CCA loss averages their cosine distance:
\begin{equation}
\mathcal L_{\mathrm{CCA}}=\mathbb E\!\left[\frac{1}{P}\sum_{p=1}^{P}
\left(1-\langle f_p,f_p^*\rangle\right)\right].
\label{eq:tf}
\end{equation}
This objective provides paired continuation supervision under reference histories. Because these histories do not contain accumulated prediction errors, we complement CCA with distribution matching on self-generated rollouts.

\subsection{Rollout Distribution Matching}
\label{sec:mmd}
\paragraph{Learning from the generated population.}
We match the training distribution $\mathcal P_\theta$ of generated video--prompt pairs to the fixed reference distribution $\mathcal Q_{\mathcal D}$. $\mathcal P_\theta$ is induced by prompt sampling, noise, and a shared random denoising exit. Each chunk is generated from noise using the key--value (KV) cache $H^{j-1}$ of earlier generated chunks. For $K=4$ denoising evaluations, the exit is shared across the rollout:
\begin{equation}
\begin{aligned}
S&\sim\operatorname{Unif}\{1,\ldots,K\},\\
\hat x^j&=F_\theta\!\left(\sg(x_S^j),\sigma_S;\sg(H^{j-1}),c\right).
\end{aligned}
\label{eq:sf-exit}
\end{equation}
Here $x_S^j$ is obtained by denoising and re-noising before exit $S$; $\sg$ detaches both this trajectory and the cached history~\cite{selfforcing,virdm}. The predictions form a complete video $\hat x=(\hat x^1,\ldots,\hat x^J)$ for distribution matching. At inference, every chunk uses all four denoising evaluations.

\begin{table*}[t]
\centering
\normalsize
\setlength{\tabcolsep}{3.5pt}
\caption{\textbf{Text-to-video generation on VBench.} Models are grouped by generation scheme. Scores are percentages. The highlighted row is our 250-update model. $^{*}$ViRDM denotes its reported variant without optical-flow regularization. Our main model also excludes this auxiliary objective; Appendix~\ref{sec:flow-ablation} presents our optical-flow regularization ablation.}
\label{tab:main}
\begin{tabularx}{\textwidth}{>{\raggedright\arraybackslash}Xccccrrr}
\toprule
\rowcolor{tablelavender}
\textbf{Model} & \textbf{\# Params} & \textbf{Resolution} & \textbf{Throughput} $\uparrow$ & \textbf{Latency (s)} $\downarrow$ & \textbf{Total} $\uparrow$ & \textbf{Quality} $\uparrow$ & \textbf{Semantic} $\uparrow$ \\
\midrule
\rowcolor{grouplight}
\multicolumn{8}{l}{\textit{Bidirectional Video Diffusion Models}} \\
LTX-Video-1.9B~\cite{ltxvideo} & 1.9B & $768\times512$ & 8.98 & 13.5 & 79.83 & 81.88 & 71.62 \\
Wan2.1-1.3B~\cite{wan} & 1.3B & $832\times480$ & 0.78 & 103 & 83.37 & 84.30 & 79.65 \\
\midrule
\rowcolor{grouplight}
\multicolumn{8}{l}{\textit{Autoregressive Video Diffusion Models}} \\
NOVA~\cite{nova} & 0.6B & $768\times480$ & 0.88 & 4.1 & 80.31 & 80.66 & 78.92 \\
Pyramid Flow~\cite{pyramidflow} & 2.0B & $640\times384$ & 6.7 & 2.5 & 80.75 & 83.41 & 70.11 \\
SkyReels-V2-1.3B~\cite{skyreelsv2} & 1.3B & $960\times540$ & 0.49 & 112 & 81.97 & 83.96 & 74.01 \\
MAGI-1-4.5B~\cite{magi} & 4.5B & $832\times480$ & 0.19 & 282 & 78.88 & 81.67 & 67.72 \\
\midrule
\rowcolor{grouplight}
\multicolumn{8}{l}{\textit{Four-Step Causal Video Models}} \\
CausVid~\cite{causvid} & 1.3B & $832\times480$ & 20.63 & 0.25 & 81.33 & 83.98 & 70.72 \\
Self Forcing~\cite{selfforcing} & 1.3B & $832\times480$ & 20.63 & 0.25 & 83.74 & 84.48 & 80.77 \\
Causal Forcing~\cite{causalforcing} & 1.3B & $832\times480$ & 20.63 & 0.25 & 84.04 & 84.59 & \textbf{81.84} \\
LongLive~\cite{longlive} & 1.3B & $832\times480$ & 20.63 & 0.25 & 83.62 & 84.36 & 80.69 \\
ViRDM$^{*}$~\cite{virdm} & 1.3B & $832\times480$ & 20.63 & 0.25 & 83.41 & 84.11 & 80.61 \\
Elastic Forcing~\cite{elastic} & 1.3B & $832\times480$ & 20.63 & 0.25 & 84.25 & 85.06 & 80.99 \\
Causal-rCM~\cite{zheng2026causal} & 1.3B & $832\times480$ & 20.63 & 0.25 & 84.37 & 85.02 & 81.73 \\
\rowcolor{black!4}
\textbf{\method (ours)} & 1.3B & $832\times480$ & 20.63 & 0.25 & \textbf{84.63} & \textbf{86.01} & 79.08 \\
\bottomrule
\end{tabularx}
\end{table*}

\paragraph{A joint video--text representation.}
For a frozen spatiotemporal encoder $e$, let $\Phi_e(x,c)\in\mathbb R^{P_e\times d_e}$ denote its full token sequence. We select fixed, ordered positions $\mathcal I_e$ of size $K_e$ and define
\begin{equation}
z_e(x,c)=\frac{1}{\sqrt{K_e}}\operatorname{vec}\!\left(\Phi_e(x,c)_{\mathcal I_e}\right).
\label{eq:tokens}
\end{equation}
Squared Euclidean distance between these representations averages squared feature differences at corresponding token positions; each complete video remains one distributional sample. To associate it with the prompt, we use a unit-normalized embedding $q(c)$ from SigLIP2~\cite{siglip2}. For $a=(x,c)$ and $a'=(x',c')$, visual and text bandwidths $s_e$ and $s_c$ define a scaled joint representation and its Gaussian kernel:
\begin{equation}
\begin{aligned}
h_e(a)&=[z_e(a);(s_e/s_c)q(c)],\\
k_e(a,a')&=\exp\!\left[-\frac{\|h_e(a)-h_e(a')\|_2^2}{2s_e^2}\right].
\end{aligned}
\label{eq:kernel}
\end{equation}
The kernel factors into visual and text similarities with bandwidths $s_e$ and $s_c$. Prompt similarity therefore weights video comparisons to match the joint video--prompt distribution.

Both bandwidths use the median heuristic on their respective reference features~\mbox{\cite{rdm,virdm}} and remain fixed during training; Appendix~\ref{sec:prompt-representations} gives the estimation details.

\paragraph{Population alignment through MMD.}
We measure the discrepancy between generated and reference populations using maximum mean discrepancy~\cite{gretton2012,rdm}:
\begin{align}
\mathcal D_e(\theta)
&=\mmd^2_{k_e}(\mathcal P_\theta,\mathcal Q_{\mathcal D})\nonumber\\
&=\mathbb E_{a,a'\sim \mathcal P_\theta}k_e(a,a')
-2\mathbb E_{a\sim \mathcal P_\theta,b\sim \mathcal Q_{\mathcal D}}k_e(a,b)\nonumber\\
&\quad+\mathbb E_{b,b'\sim \mathcal Q_{\mathcal D}}k_e(b,b').
\label{eq:mmd}
\end{align}
All paired draws within an expectation are independent. Generated--reference interactions attract the population toward the reference kernel mean, while generated--generated interactions discourage concentration in representation space. The target is sample-defined: no diffusion-score teacher or learned fake-score model is required.

For optimization, we normalize each encoder's empirical discrepancy by its stop-gradient value, then average these normalized objectives equally to obtain $\mathcal L_{\mathrm{RDM}}$; Appendix~\ref{sec:empirical-objective} specifies the estimator and normalization. We select the representation spaces for continuation alignment and rollout distribution matching independently.

\subsection{Dynamic Gradient Balancing}
\label{sec:controller}
\begingroup
\setlength{\abovedisplayskip}{6pt plus 2pt minus 2pt}
\setlength{\belowdisplayskip}{6pt plus 2pt minus 2pt}
The two objectives may produce gradients with different magnitudes, making fixed loss weights sensitive to their raw scales. We therefore separate gradient norm calibration~\cite{gradnorm} from error-dependent weighting: gradient norm matching establishes a common scale, while residual error weighting adjusts the relative contribution of continuation alignment.

Let $g_t^{\mathrm{CCA}}=\nabla_\theta\mathcal L_{\mathrm{CCA},t}$ and $g_t^{\mathrm{RDM}}=\nabla_\theta\mathcal L_{\mathrm{RDM},t}$. \emph{Gradient norm matching (GNM)} first rescales the continuation alignment gradient to the RDM gradient norm:
\begin{equation}
\bar g_t^{\mathrm{CCA}}=
\|g_t^{\mathrm{RDM}}\|_2\frac{g_t^{\mathrm{CCA}}}{\|g_t^{\mathrm{CCA}}\|_2}.
\label{eq:calibration}
\end{equation}
This preserves the continuation gradient direction while matching its scale to distribution matching.

\emph{Residual error weighting (REW)} then uses the current fitting error to set the CCA-to-RDM gradient norm ratio before clipping. As reference continuations become better fitted, we relax the relative constraint of paired supervision so rollout distribution learning plays a larger role, while retaining a lower bound on the continuation contribution. Let $\ell_t$ denote the current-batch CCA loss and $\ell_1$ its value at the first update. We set
\begin{equation}
\begin{aligned}
r_t&=r_{\min}+(1-r_{\min})\min\!\left(\frac{\ell_t}{\ell_1},1\right),\\
g_t&=g_t^{\mathrm{RDM}}+r_t\bar g_t^{\mathrm{CCA}}.
\end{aligned}
\label{eq:ratio}
\end{equation}
where $r_{\min}=0.1$. The weight $r_t$ follows batch-to-batch error fluctuations and can increase or decrease during training.
\par\endgroup

\section{Experiments}
\label{sec:experiments}
\setlength{\textfloatsep}{10pt plus 2pt minus 2pt}
\setlength{\floatsep}{6pt plus 2pt minus 2pt}
\setlength{\intextsep}{8pt plus 2pt minus 2pt}
\raggedbottom
\subsection{Experimental Setup}
\label{sec:experimental-setup}
\paragraph{Model and training.}
We initialize \method from pretrained bidirectional Wan2.1-T2V-1.3B~\cite{wan} and generate 81-frame (21-latent-frame) videos at $832\times480$ resolution in autoregressive chunks of three latent frames. Unless otherwise specified, training and inference use four denoising steps with timestep shift 5. We perform 250 joint updates on eight NVIDIA B200 GPUs using 6,505 reference video--prompt pairs released by Causal Forcing~\cite{causalforcing}. Each update uses 128 generated videos for RDM and 32 reference pairs for CCA. Our main configuration uses \vm~\cite{videomaev2} L24 for CCA and \vm L16 plus DiT L9 for RDM, with frozen SigLIP2~\cite{siglip2} prompt embeddings for the MMD kernel. The generator retains its native text encoder. DiT features come from frozen Wan2.1-T2V-14B~\cite{wan}. Reference features and prompt embeddings are precomputed offline. Appendices~\ref{sec:implementation}--\ref{sec:estimation} detail the implementation; Appendix~\ref{sec:dit-feature-studies} examines DiT feature choices.

\paragraph{Evaluation.}
We evaluate \method on the 944 expanded VBench~\cite{vbench} prompts released by Causal Forcing~\cite{causalforcing}, generating five videos per prompt (4,720 videos in total) at 16 frames/s. Final videos are decoded with the original Wan VAE. Baseline Total, Quality, and Semantic scores follow published reports, using ViRDM's variant without optical-flow regularization and Elastic Forcing's three-encoder configuration (Appendix~\ref{sec:baseline-score-sources}). Throughput (frames/s) and latency for the four-step causal models are measured on a single NVIDIA B200 GPU; other timing values follow published reports. Latency denotes time to the first frame. Bold marks the best score for each metric unless otherwise stated.

\begin{figure}[t]
\centering
\includegraphics[width=\columnwidth]{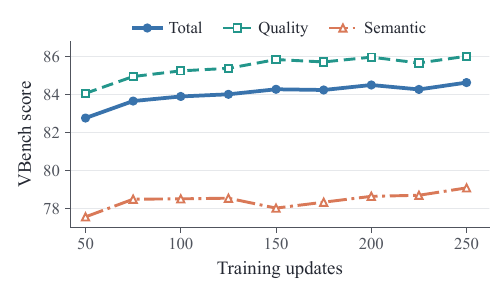}
\caption{\textbf{Effect of training duration.} VBench scores of our main configuration (RDM batch size 128) from 50 to 250 updates. Each marker denotes an evaluated checkpoint.}
\label{fig:training-duration}
\end{figure}

\subsection{Main Results}
\method achieves 84.63 Total and 86.01 Quality with four denoising evaluations per chunk (Table~\ref{tab:main}), giving the highest Total and Quality among the listed four-step causal models. Compared with Elastic Forcing, \method improves Total by 0.38 and Quality by 0.95, while Semantic is 1.91 points lower. \method also achieves a Dynamic Degree of 75.56, compared with 63.33 for Elastic Forcing and 48.61 reported for ViRDM (Appendix~\ref{sec:main-dynamic}).

\method reaches this performance through 250 joint updates directly from the bidirectional backbone, without a separate initialization stage. Figure~\ref{fig:training-duration} shows that Total already reaches 84.27 after 150 updates and improves overall to 84.63 at 250 updates. The chunk-boundary ratio compares frame differences at boundaries and within chunks, normalized by the corresponding reference-video ratio. Our ratio is closer to one than ViRDM's (1.040 vs.\ 1.393; Appendix~\ref{sec:temporal-continuity}), supporting temporal continuity without a separate initialization stage.

\method learns from closed-source video outputs without access to the reference generator's weights. Using curated high-quality Seedance 2.5~\cite{seedance25} motion videos raises Dynamic Degree from 54.72 to 62.50 (Appendix~\ref{sec:closed-source-data}).

\subsection{Ablation Studies}
\label{sec:ablations}
Unless otherwise specified, ablations use 150 updates with SigLIP2 prompt embeddings and RDM and CCA batch sizes of 64 and 32, respectively. The component ablation (including joint training) and the gradient-balancing, batch-size, and prompt-representation ablations use \vm L16 for each active component. For the representation studies, we first vary each component's representation while fixing the other to \vm L16, then evaluate combinations of the selected representations.

\begin{table}[t]
\centering
\captionsetup{skip=4pt}
\footnotesize
\setlength{\tabcolsep}{2pt}
\renewcommand{\arraystretch}{1.12}
\caption{\textbf{CCA and RDM training.} Subj./Smooth./Aes./Dyn.: Subject Consistency/Motion Smoothness/Aesthetic Quality/Dynamic Degree (VBench quality); Sem.: aggregate Semantic. Sequential uses $150+150$ updates; others use 150.}
\label{tab:component-ablation}
\begin{tabularx}{\columnwidth}{@{}>{\raggedright\arraybackslash}Xrrrrr@{}}
\toprule
\rowcolor{tablelavender}
\textbf{Configuration} & \textbf{Subj.} $\uparrow$ & \textbf{Smooth.} $\uparrow$ & \textbf{Aes.} $\uparrow$ & \textbf{Dyn.} $\uparrow$ & \textbf{Sem.} $\uparrow$ \\
\midrule
CCA only & \textbf{97.08} & \textbf{99.50} & 59.13 & 13.89 & 72.21 \\
RDM only & 84.17 & 96.82 & 63.95 & 11.11 & 71.93 \\
CCA init $\rightarrow$ RDM-only & 94.08 & 98.56 & 64.35 & \textbf{74.44} & 77.88 \\
\rowcolor{black!4}
CCA + RDM jointly & 95.69 & 98.77 & \textbf{65.07} & 54.72 & \textbf{77.99} \\
\bottomrule
\end{tabularx}
\end{table}

\paragraph{Joint training of CCA and RDM.}
Table~\ref{tab:component-ablation} shows that CCA-only achieves the highest Subject Consistency and Motion Smoothness but yields low Dynamic and degraded self-rollouts, whereas RDM-only improves Aesthetic at the expense of consistency. Joint training achieves the highest Aesthetic and Semantic scores while retaining high consistency. Sequential training yields higher Dynamic (74.44 vs.\ 54.72) but lower scores on all other reported metrics, despite twice as many updates. Its transient distortions and artifacts (Appendix~\ref{sec:component-ablation}) further support maintaining continuation alignment throughout rollout distribution matching.

\begin{figure}[t]
\centering
\includegraphics[width=\columnwidth]{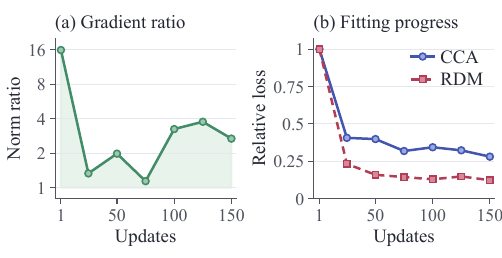}
\caption{\textbf{Gradient scales and fitting progress.} (a) Fixed-weight CCA/RDM gradient norm ratio. (b) GNM-only losses relative to update 1; RDM is measured before self-normalization.}
\label{fig:gradient-scale}
\par\smallskip
\begingroup
\captionsetup{type=table}
\begingroup
\centering
\footnotesize
\setlength{\tabcolsep}{1.7pt}
\renewcommand{\arraystretch}{1.16}
\caption{Dynamic Gradient Balancing with both objectives. Fixed weights omit GNM; $r$ is the fixed post-GNM CCA/RDM gradient norm ratio.}
\label{tab:controller}
\begin{tabularx}{\columnwidth}{@{}>{\raggedright\arraybackslash}p{.26\columnwidth}*{4}{>{\raggedleft\arraybackslash}X}@{}}
\toprule
\rowcolor{tablelavender}
\textbf{Configuration} & \textbf{Total} $\uparrow$ & \textbf{Quality} $\uparrow$ & \textbf{Semantic} $\uparrow$ & \textbf{Dynamic} $\uparrow$ \\
\midrule
Fixed weights & 80.43 & 81.29 & 77.00 & 31.11 \\
GNM ($r=1$) & 81.23 & 82.04 & \textbf{78.01} & 37.22 \\
GNM ($r=0.5$) & 81.79 & 82.78 & 77.80 & 45.83 \\
\rowcolor{black!4}
GNM + REW & \textbf{82.40} & \textbf{83.51} & 77.99 & \textbf{54.72} \\
\bottomrule
\end{tabularx}
\par\endgroup

\endgroup
\end{figure}

\paragraph{Dynamic gradient balancing.}
CCA gradients dominate under fixed weights, with an initial CCA/RDM norm ratio of 15.9 (Figure~\ref{fig:gradient-scale}(a)). GNM raises Total from 80.43 to 81.23 and Dynamic from 31.11 to 37.22 (Table~\ref{tab:controller}). Yet CCA loss falls to 28\% of its initial value by update 150 (Figure~\ref{fig:gradient-scale}(b)), motivating residual-based weighting. A fixed post-GNM ratio $r=0.5$ yields 81.79 Total and 45.83 Dynamic. REW improves these to 82.40 and 54.72, outperforming this control on all four metrics and supporting adaptive weighting beyond norm calibration or a fixed reduction.

\begin{table}[t]
\captionsetup{position=top,skip=10pt}
\begingroup
\centering
\footnotesize
\setlength{\tabcolsep}{2pt}
\renewcommand{\arraystretch}{1.16}
\caption{RDM batch size with VideoMAE v2 L16 for both components. CCA batch size is 32; all runs use 150 updates.}
\label{tab:rdm-batch-size}
\begin{tabularx}{\linewidth}{@{}l*{6}{>{\centering\arraybackslash}X}@{}}
\toprule
\rowcolor{tablelavender}
\textbf{Batch size} & \textbf{32} & \textbf{64} & \textbf{128} & \textbf{256} & \textbf{512} & \textbf{1024} \\
\midrule
Total $\uparrow$ & 81.27 & 82.40 & 82.36 & \textbf{82.96} & 82.89 & 82.81 \\
Quality $\uparrow$ & 82.10 & 83.51 & 83.46 & 83.95 & \textbf{84.06} & 83.89 \\
Semantic $\uparrow$ & 77.95 & 77.99 & 77.99 & \textbf{78.97} & 78.21 & 78.47 \\
Dynamic $\uparrow$ & 33.33 & \textbf{54.72} & 47.22 & 47.78 & 54.17 & 51.67 \\
\bottomrule
\end{tabularx}
\par
\endgroup

\par\vspace{18pt}
\begingroup
\centering
\footnotesize
\setlength{\tabcolsep}{1.7pt}
\renewcommand{\arraystretch}{1.16}
\caption{Prompt encoder in the MMD kernel. Both components use VideoMAE v2 L16.}
\label{tab:text}
\begin{tabularx}{\linewidth}{@{}>{\raggedright\arraybackslash}Xrrrr@{}}
\toprule
\rowcolor{tablelavender}
\textbf{Prompt encoder} & \textbf{Total} $\uparrow$ & \textbf{Quality} $\uparrow$ & \textbf{Semantic} $\uparrow$ & \textbf{Dynamic} $\uparrow$ \\
\midrule
None & 81.60 & 82.55 & 77.81 & 39.72 \\
Qwen3-VL-Embedding & 81.97 & 83.07 & 77.53 & 45.56 \\
\rowcolor{black!4}
SigLIP2 & \textbf{82.40} & \textbf{83.51} & \textbf{77.99} & \textbf{54.72} \\
\bottomrule
\end{tabularx}
\par
\endgroup

\end{table}

\paragraph{RDM batch size.}
Table~\ref{tab:rdm-batch-size} varies the global RDM batch size under the common ablation configuration. Larger batches generally improve Total, though not monotonically, with a peak of 82.96 at batch size 256. Appendix~\ref{sec:main-rdm-batch-size} reports batch-size selection for the main model.

\paragraph{Prompt representation.}
Table~\ref{tab:text} varies the prompt representation used in the MMD kernel. Compared with visual-only matching, SigLIP2 raises Total from 81.60 to 82.40 and Dynamic from 39.72 to 54.72. It also outperforms \mbox{Qwen3-VL-Embedding-2B}~\cite{qwenembedding} on all four metrics. These results support incorporating prompt similarity into population matching and motivate SigLIP2 as the default prompt representation in our experiments.

\paragraph{Representation for causal continuation alignment.}
\mbox{Table~\ref{tab:tf-space}} compares representations for continuation alignment, with rollout distribution matching fixed to \vm L16. For both VideoMAE v2 and V-JEPA 2.1, L24 yields higher scores than L16 on all four metrics. \vm L24 achieves the highest Total and Quality, exceeding latent FM by 1.04 and 1.23 points, respectively, while DiT L9 gives the highest Semantic and Dynamic. We therefore select \vm L24 for continuation alignment.

\begin{table}[t]
\captionsetup{position=top,skip=5pt,justification=raggedright,singlelinecheck=false,margin=0pt}
\begingroup
\centering
\footnotesize
\setlength{\tabcolsep}{1.7pt}
\renewcommand{\arraystretch}{1.16}
\caption{Representation for causal continuation alignment, with rollout distribution matching fixed to VideoMAE v2 L16. Latent FM denotes velocity regression in latent space.}
\label{tab:tf-space}
\begin{tabularx}{\linewidth}{@{}>{\raggedright\arraybackslash}Xrrrr@{}}
\toprule
\rowcolor{tablelavender}
\textbf{Representation} & \textbf{Total} $\uparrow$ & \textbf{Quality} $\uparrow$ & \textbf{Semantic} $\uparrow$ & \textbf{Dynamic} $\uparrow$ \\
\midrule
Latent FM & 81.76 & 82.77 & 77.75 & 51.39 \\
\addlinespace[3pt]
V-JEPA 2.1 L16 & 81.77 & 83.12 & 76.35 & 40.56 \\
V-JEPA 2.1 L24 & 82.05 & 83.30 & 77.08 & 58.06 \\
\addlinespace[3pt]
DiT L9 & 82.56 & 83.53 & \textbf{78.68} & \textbf{58.89} \\
\addlinespace[3pt]
VideoMAE v2 L16 & 82.40 & 83.51 & 77.99 & 54.72 \\
\rowcolor{black!4}
VideoMAE v2 L24 & \textbf{82.80} & \textbf{84.00} & 78.01 & 58.06 \\
\bottomrule
\end{tabularx}
\par
\endgroup

\par\vspace{9pt}
\begingroup
\centering
\footnotesize
\setlength{\tabcolsep}{1.7pt}
\renewcommand{\arraystretch}{1.16}
\caption{Representation for rollout distribution matching, with continuation alignment fixed to VideoMAE v2 L16.}
\label{tab:mmd-space}
\begin{tabularx}{\linewidth}{@{}>{\raggedright\arraybackslash}Xrrrr@{}}
\toprule
\rowcolor{tablelavender}
\textbf{Representation} & \textbf{Total} $\uparrow$ & \textbf{Quality} $\uparrow$ & \textbf{Semantic} $\uparrow$ & \textbf{Dynamic} $\uparrow$ \\
\midrule
VideoMAE v2 L16 & 82.40 & 83.51 & 77.99 & \textbf{54.72} \\
VideoMAE v2 L24 & 80.75 & 81.42 & 78.07 & 28.33 \\
\addlinespace[3pt]
V-JEPA 2.1 L16 & 80.51 & 81.67 & 75.86 & 15.56 \\
V-JEPA 2.1 L24 & 79.88 & 80.82 & 76.10 & 8.06 \\
\addlinespace[3pt]
\rowcolor{black!4}
DiT L9 & \textbf{82.45} & \textbf{83.53} & \textbf{78.13} & 52.22 \\
\bottomrule
\end{tabularx}
\par
\endgroup

\end{table}

\paragraph{Representation for rollout distribution matching.}
\mbox{Table~\ref{tab:mmd-space}} compares representations for rollout distribution matching, with continuation alignment fixed to \vm L16. For both VideoMAE v2 and V-JEPA 2.1, L16 yields higher Total, Quality, and Dynamic than L24, while L24 gives slightly higher Semantic. For \vm, Dynamic drops from 54.72 at L16 to 28.33 at L24. DiT L9 achieves the highest Total and Semantic, whereas \vm L16 gives the highest Dynamic. The opposite layer preferences across the two components within these encoders support selecting representations separately for continuation alignment and distribution matching.

\begin{table}[t]
\vspace{3pt}
\captionsetup{skip=6pt}
\begingroup
\centering
\footnotesize
\setlength{\tabcolsep}{1.5pt}
\renewcommand{\arraystretch}{1.16}
\caption{Representation combinations after 150 updates. $V_k$: VideoMAE v2 L$k$; $D_9$: DiT L9. Gray marks the selected combination.}
\label{tab:representation-combinations}
\begin{tabularx}{\linewidth}{*{2}{>{\centering\arraybackslash}X}rrrr}
\toprule
\rowcolor{tablelavender}
\textbf{Continuation} & \textbf{Distribution} & \textbf{Total} $\uparrow$ & \textbf{Qual.} $\uparrow$ & \textbf{Sem.} $\uparrow$ & \textbf{Dyn.} $\uparrow$ \\
\midrule
$V_{24}$ & $V_{16}$ & 82.80 & 84.00 & 78.01 & 58.06 \\
$V_{24}$ & $D_{9}$ & 83.49 & 84.96 & 77.58 & 68.61 \\
\rowcolor{black!4}
$V_{24}$ & $V_{16}+D_{9}$ & \textbf{84.11} & \textbf{85.47} & \textbf{78.67} & 74.72 \\
$V_{24}+D_{9}$ & $V_{16}+D_{9}$ & 83.68 & 84.99 & 78.43 & 71.94 \\
$V_{24}+D_{9}$ & $V_{16}+D_{9}+V_{24}$ & 83.46 & 84.71 & 78.45 & \textbf{75.00} \\
\bottomrule
\end{tabularx}
\par
\endgroup

\end{table}

\begin{figure*}[t]
\centering
\captionsetup{skip=7pt}
\includegraphics[width=0.95\textwidth]{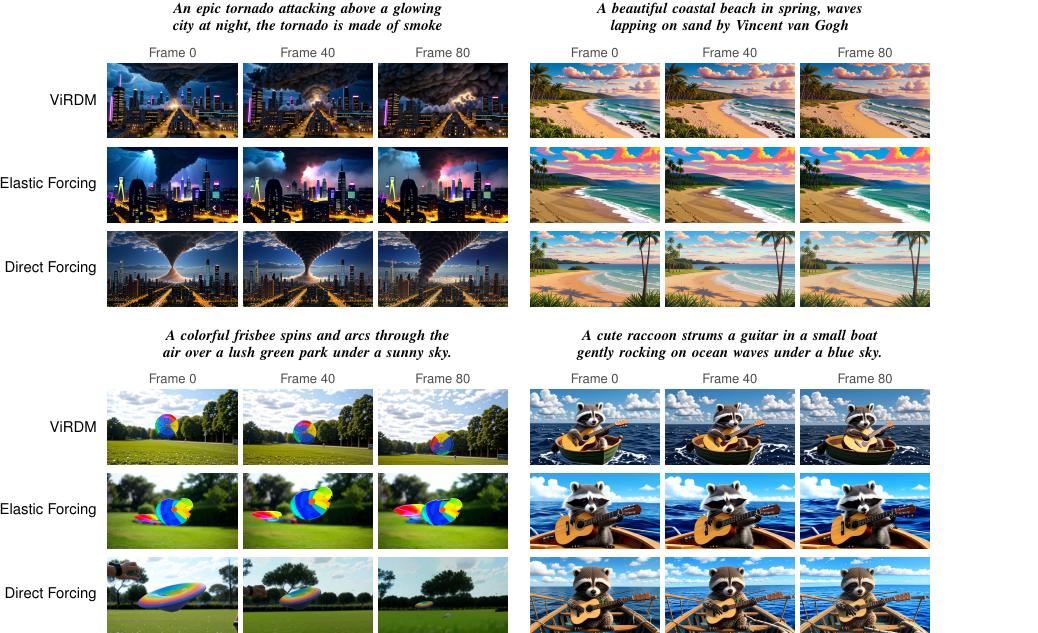}
\caption{\textbf{Qualitative comparison on four text-to-video prompts.} For each case, rows show ViRDM, Elastic Forcing, and \method from top to bottom; columns show frames 0, 40, and 80. These lengthy prompts are abbreviated above the panels for readability.}
\label{fig:qualitative}
\end{figure*}

\paragraph{Combining representation spaces.}
Table~\ref{tab:representation-combinations} evaluates combinations of the top two representations by Total for each component identified above. With CCA fixed to $V_{24}$, combining $V_{16}$ and $D_9$ for RDM improves all four metrics over either representation alone, raising Total to 84.11 from 82.80 and 83.49, respectively. With this RDM configuration fixed, adding $D_9$ to CCA instead reduces Total to 83.68. We therefore adopt $V_{24}$ for CCA and $V_{16}+D_9$ for RDM, which achieves the highest Total, Quality, and Semantic among the tested combinations. All feature encoders are used only during training and add no inference overhead.

\subsection{Qualitative Comparisons}
\label{sec:qualitative}
Figure~\ref{fig:qualitative} compares ViRDM, Elastic Forcing, and \method under the same prompts. The tornado and frisbee examples highlight structural preservation: \method retains a distinct funnel and a disc-shaped object across the displayed frames, while the baseline frisbee results exhibit umbrella-like or folded forms. The coastal and raccoon examples highlight texture and local detail, with \method preserving the structure of clouds and waves, the texture of the raccoon's fur, and fine details of the guitar.

\raggedbottom
\setlength{\textfloatsep}{8pt}
\begin{table}[t]
\centering
\captionsetup{skip=4pt}
\small
\setlength{\tabcolsep}{3pt}
\renewcommand{\arraystretch}{1.16}
\caption{Long-video generation on MovieGen-128 at 50 seconds. Self-Forcing++ scores follow its reported results~\cite{cui2026self}. All scores are percentages; higher is better.}
\label{tab:long-video}
\begin{tabularx}{\columnwidth}{@{}l>{\centering\arraybackslash}X>{\columncolor{black!4}\centering\arraybackslash}X@{}}
\toprule
\rowcolor{tablelavender}
\textbf{Metric} & \textbf{Self-Forcing++} & \textbf{\methodname} \\
\midrule
Subject consistency & 97.00 & \textbf{97.07} \\
Background consistency & 95.55 & \textbf{96.11} \\
Motion smoothness & 98.39 & \textbf{98.61} \\
Dynamic degree & \textbf{55.36} & 53.13 \\
Aesthetic quality & 53.76 & \textbf{56.14} \\
Imaging quality & \textbf{67.87} & 67.15 \\
\bottomrule
\end{tabularx}
\end{table}

\subsection{Long-Video Generation}
\label{sec:long-video}
We adapt the 150-update short-video checkpoint with 100 further updates using only the original 6,505 short reference videos. Following window-based long-rollout training~\cite{cui2026self}, RDM supervises approximately five-second windows sampled from self-generated rollouts of up to approximately 60 seconds. We match their globally pooled \vm L16 and DiT L9 features to the short-reference distribution, while retaining token-level CCA with \vm L24 throughout adaptation. Thus, adaptation learns from long generated histories without long references or an online diffusion teacher. Evaluation uses 50-second videos generated from MovieGen-128 prompts.

\method outperforms Self-Forcing++~\cite{cui2026self} on four of six VBench-Long~\cite{vbenchlong} metrics, with lower Dynamic Degree and Imaging Quality (Table~\ref{tab:long-video}; details in Appendix~\ref{sec:long-video-details}).

\begingroup
\brokenpenalty=10000
\section{Conclusion}
We presented \method, a single-stage framework that directly adapts a pretrained bidirectional video model into a four-step autoregressive generator in 250 joint updates, without a separate initialization stage. Causal Continuation Alignment learns paired continuations under reference histories, while Rollout Distribution Matching aligns self-generated rollouts with the reference population. Dynamic Gradient Balancing controls their relative gradient contributions as continuation fitting progresses. Our ablations support both gradient-balancing components and objective-specific representation choices, including the benefit of combining features for rollout distribution matching. The resulting model achieves 84.63 VBench Total and 86.01 Quality. Short references also support 50-second generation.

\paragraph{Limitations.}
Our evaluation covers a 1.3B generator at $832\times480$ resolution and durations up to 50 seconds. Testing larger models, higher resolutions, and longer generation horizons remains an important direction for future work.

\par
\endgroup
\flushbottom
\setlength{\textfloatsep}{20pt plus 2pt minus 4pt}
\setlength{\floatsep}{12pt plus 2pt minus 2pt}
\setlength{\intextsep}{12pt plus 2pt minus 2pt}
{\small
\bibliographystyle{ieeenat_fullname}
\bibliography{main}}

@article{virdm,
 author={Meng, Zichong and Ge, Chongjian and Huang, Chun-Hao P. and Zhou, Yang and Jiang, Huaizu},
 title={{ViRDM}: Taming Representation Distribution Matching for Few-Step Causal Video Generation},
 journal={arXiv preprint arXiv:2609.28923}, year={2026}
}

@article{elastic,
 author={Zhang, Chi and Liu, Yueyi and Shi, Haoyang and An, Ruichuan and Li, Haoyu and Wu, Yuhang and Cui, Sen and Liu, Miao},
 title={From Scores to Samples: Elastic Forcing for Autoregressive Video Generation},
 journal={arXiv preprint arXiv:2609.35491}, year={2026}
}

@inproceedings{selfforcing,
 author={Huang, Xun and Li, Zhengqi and He, Guande and Zhou, Mingyuan and Shechtman, Eli},
 title={Self Forcing: Bridging the Train-Test Gap in Autoregressive Video Diffusion},
 booktitle={NeurIPS}, year={2025}
}

@article{causalforcing,
 author={Zhu, Hongzhou and Zhao, Min and He, Guande and Su, Hang and Li, Chongxuan and Zhu, Jun},
 title={Causal Forcing: Autoregressive Diffusion Distillation Done Right for High-Quality Real-Time Interactive Video Generation},
 journal={arXiv preprint arXiv:2602.02214}, year={2026}
}

@inproceedings{causvid,
 author={Yin, Tianwei and Zhang, Qiang and Zhang, Richard and Freeman, William T. and Durand, Fredo and Shechtman, Eli and Huang, Xun},
 title={From Slow Bidirectional to Fast Autoregressive Video Diffusion Models},
 booktitle={CVPR}, year={2025}, pages={22963--22974}
}

@article{wan,
 author={{Wan Team} and Wang, Ang and Ai, Baole and Wen, Bin and Mao, Chaojie and Xie, Chen-Wei and others},
 title={Wan: Open and Advanced Large-Scale Video Generative Models},
 journal={arXiv preprint arXiv:2503.20314}, year={2025}
}

@inproceedings{dmd,
 author={Yin, Tianwei and Gharbi, Micha\"el and Zhang, Richard and Shechtman, Eli and Durand, Fredo and Freeman, William T. and Park, Taesung},
 title={One-Step Diffusion with Distribution Matching Distillation},
 booktitle={CVPR}, year={2024}
}

@inproceedings{dmd2,
 author={Yin, Tianwei and Gharbi, Micha\"el and Park, Taesung and Zhang, Richard and Shechtman, Eli and Durand, Fredo and Freeman, William T.},
 title={Improved Distribution Matching Distillation for Fast Image Synthesis},
 booktitle={NeurIPS}, year={2024}
}

@article{gretton2012,
 author={Gretton, Arthur and Borgwardt, Karsten M. and Rasch, Malte J. and Sch\"olkopf, Bernhard and Smola, Alexander},
 title={A Kernel Two-Sample Test}, journal={Journal of Machine Learning Research}, volume={13}, pages={723--773}, year={2012}
}

@article{rdm,
 author={Feng, Lan and Li, Wuyang and Zablocki, {\'E}loi and Cord, Matthieu and Alahi, Alexandre},
 title={Representation Distribution Matching for One-Step Visual Generation},
 journal={arXiv preprint arXiv:2607.02375}, year={2026}
}

@inproceedings{videomaev2,
 author={Wang, Limin and Huang, Bingkun and Zhao, Zhiyu and Tong, Zhan and He, Yinan and Wang, Yi and Wang, Yali and Qiao, Yu},
 title={{VideoMAE V2}: Scaling Video Masked Autoencoders with Dual Masking},
 booktitle={CVPR}, pages={14549--14560}, year={2023}
}

@article{vjepa21,
 author={Mur-Labadia, Lorenzo and Muckley, Matthew and Bar, Amir and Assran, Mido and Sinha, Koustuv and Rabbat, Mike and LeCun, Yann and Ballas, Nicolas and Bardes, Adrien},
 title={{V-JEPA 2.1}: Unlocking Dense Features in Video Self-Supervised Learning},
 journal={arXiv preprint arXiv:2603.14482}, year={2026}
}

@inproceedings{vbench,
 author={Huang, Ziqi and He, Yinan and Yu, Jiashuo and Zhang, Fan and Si, Chenyang and Jiang, Yuming and Zhang, Yuanhan and Wu, Tianxing and Jin, Qingyang and Chanpaisit, Nattapol and Wang, Yaohui and Chen, Xinyuan and Wang, Limin and Lin, Dahua and Qiao, Yu and Liu, Ziwei},
 title={{VBench}: Comprehensive Benchmark Suite for Video Generative Models},
 booktitle={CVPR}, year={2024}
}

@article{siglip2,
 author={Tschannen, Michael and Gritsenko, Alexey and Wang, Xiao and Naeem, Muhammad Ferjad and Alabdulmohsin, Ibrahim and Parthasarathy, Nikhil and Evans, Talfan and Beyer, Lucas and Xia, Ye and Mustafa, Basil and H\'enaff, Olivier and Harmsen, Jeremiah and Steiner, Andreas and Zhai, Xiaohua},
 title={{SigLIP 2}: Multilingual Vision-Language Encoders with Improved Semantic Understanding, Localization, and Dense Features},
 journal={arXiv preprint arXiv:2502.14786}, year={2025}
}

@article{qwenembedding,
 author={Li, Mingxin and Zhang, Yanzhao and Long, Dingkun and Chen, Keqin and Song, Sibo and Bai, Shuai and Yang, Zhibo and Xie, Pengjun and Yang, An and Liu, Dayiheng and Zhou, Jingren and Lin, Junyang},
 title={{Qwen3-VL-Embedding} and {Qwen3-VL-Reranker}: A Unified Framework for State-of-the-Art Multimodal Retrieval and Ranking},
 journal={arXiv preprint arXiv:2601.04720}, year={2026}
}

@inproceedings{gradnorm,
 author={Chen, Zhao and Badrinarayanan, Vijay and Lee, Chen-Yu and Rabinovich, Andrew},
 title={{GradNorm}: Gradient Normalization for Adaptive Loss Balancing in Deep Multitask Networks},
 booktitle={ICML}, year={2018}, pages={794--803}
}

@misc{taew,
 author={Bohan, Ollin Boer}, title={{TAEHV}: Tiny Autoencoder for Hunyuan Video},
 howpublished={\url{https://github.com/madebyollin/taehv}}, year={2025}
}

@inproceedings{longlive,
 author={Yang, Shuai and Huang, Wei and Chu, Ruihang and Xiao, Yicheng and Zhao, Yuyang and Wang, Xianbang and Li, Muyang and Xie, Enze and Chen, Yingcong and Lu, Yao and Han, Song and Chen, Yukang},
 title={{LongLive}: Real-Time Interactive Long Video Generation},
 booktitle={ICLR}, year={2026}
}

@inproceedings{diffusionforcing,
 author={Chen, Boyuan and Mart{\'\i} Mons{\'o}, Diego and Du, Yilun and Simchowitz, Max and Tedrake, Russ and Sitzmann, Vincent},
 title={Diffusion Forcing: Next-Token Prediction Meets Full-Sequence Diffusion},
 booktitle={NeurIPS},
 volume={37},
 pages={24081--24125},
 year={2024}
}

@inproceedings{pyramidflow,
 author={Jin, Yang and Sun, Zhicheng and Li, Ningyuan and Xu, Kun and Jiang, Hao and Zhuang, Nan and Huang, Quzhe and Song, Yang and Mu, Yadong and Lin, Zhouchen},
 title={Pyramidal Flow Matching for Efficient Video Generative Modeling},
 booktitle={ICLR},
 pages={23378--23402},
 year={2025}
}

@article{magi,
 author={Teng, Hansi and Jia, Hongyu and Sun, Lei and Li, Lingzhi and Li, Maolin and Tang, Mingqiu and Han, Shuai and Zhang, Tianning and Zhang, WQ and Luo, Weifeng and others},
 title={{MAGI-1}: Autoregressive Video Generation at Scale},
 journal={arXiv preprint arXiv:2505.13211},
 year={2025}
}

@article{drifting,
 author={Deng, Mingyang and Li, He and Li, Tianhong and Du, Yilun and He, Kaiming},
 title={Generative Modeling via Drifting},
 journal={arXiv preprint arXiv:2602.04770},
 year={2026}
}

@article{rfl,
 author={Yang, Jiawei and Geng, Zhengyang and Ju, Xuan and Tian, Yonglong and Wang, Yue},
 title={Representation {Fr{\'e}chet} Loss for Visual Generation},
 journal={arXiv preprint arXiv:2604.28190},
 year={2026}
}

@article{wflow,
 author={Han, Jiaqi and Li, Puheng and Guo, Qiushan and Xu, Renyuan and Ermon, Stefano and Cand{\`e}s, Emmanuel J},
 title={One-Step Generative Modeling via {Wasserstein} Gradient Flows},
 journal={arXiv preprint arXiv:2605.11755},
 year={2026}
}

@article{ltxvideo,
 author={HaCohen, Yoav and Chiprut, Nisan and Brazowski, Benny and Shalem, Daniel and Moshe, Dudu and Richardson, Eitan and Levin, Eran and Shiran, Guy and Zabari, Nir and Gordon, Ori and others},
 title={{LTX-Video}: Realtime Video Latent Diffusion},
 journal={arXiv preprint arXiv:2501.00103},
 year={2024}
}

@inproceedings{nova,
 author={Deng, Haoge and Pan, Ting and Diao, Haiwen and Luo, Zhengxiong and Cui, Yufeng and Lu, Huchuan and Shan, Shiguang and Qi, Yonggang and Wang, Xinlong},
 title={Autoregressive Video Generation without Vector Quantization},
 booktitle={ICLR},
 pages={44730--44745},
 year={2025}
}

@article{skyreelsv2,
 author={Chen, Guibin and Lin, Dixuan and Yang, Jiangping and Lin, Chunze and Zhu, Junchen and Fan, Mingyuan and Zhang, Hao and Chen, Sheng and Chen, Zheng and Ma, Chengcheng and others},
 title={{SkyReels-V2}: Infinite-Length Film Generative Model},
 journal={arXiv preprint arXiv:2504.13074},
 year={2025}
}

@article{videogpt,
 author={Yan, Wilson and Zhang, Yunzhi and Abbeel, Pieter and Srinivas, Aravind},
 title={{VideoGPT}: Video Generation Using {VQ-VAE} and Transformers},
 journal={arXiv preprint arXiv:2104.10157},
 year={2021}
}

@inproceedings{videopoet,
 author={Kondratyuk, Dan and Yu, Lijun and Gu, Xiuye and Lezama, Jos{\'e} and Huang, Jonathan and Schindler, Grant and Hornung, Rachel and Birodkar, Vighnesh and Yan, Jimmy and Chiu, Ming-Chang and Somandepalli, Krishna and Akbari, Hassan and Alon, Yair and Cheng, Yong and Dillon, Joshua V. and Gupta, Agrim and Hahn, Meera and Hauth, Anja and Hendon, David and Martinez, Alonso and Minnen, David and Sirotenko, Mikhail and Sohn, Kihyuk and Yang, Xuan and Adam, Hartwig and Yang, Ming-Hsuan and Essa, Irfan and Wang, Huisheng and Ross, David A and Seybold, Bryan and Jiang, Lu},
 title={{VideoPoet}: A Large Language Model for Zero-Shot Video Generation},
 booktitle={ICML},
 volume={235},
 pages={25105--25124},
 year={2024}
}

@article{ca2vdm,
 author={Gao, Kaifeng and Shi, Jiaxin and Zhang, Hanwang and Wang, Chunping and Xiao, Jun and Chen, Long},
 title={{Ca2-VDM}: Efficient Autoregressive Video Diffusion Model with Causal Generation and Cache Sharing},
 journal={arXiv preprint arXiv:2411.16375},
 year={2024}
}

@article{historyguided,
 author={Song, Kiwhan and Chen, Boyuan and Simchowitz, Max and Du, Yilun and Tedrake, Russ and Sitzmann, Vincent},
 title={History-Guided Video Diffusion},
 journal={arXiv preprint arXiv:2502.06764},
 year={2025}
}

@inproceedings{framepack,
 author={Zhang, Lvmin and Cai, Shengqu and Li, Muyang and Wetzstein, Gordon and Agrawala, Maneesh},
 title={Frame Context Packing and Drift Prevention in Next-Frame-Prediction Video Diffusion Models},
 booktitle={NeurIPS},
 volume={38},
 pages={30546--30566},
 year={2025}
}

@inproceedings{gmmn,
 author={Li, Yujia and Swersky, Kevin and Zemel, Rich},
 title={Generative Moment Matching Networks},
 booktitle={ICML},
 volume={37},
 pages={1718--1727},
 year={2015}
}

@inproceedings{mmdnets,
 author={Dziugaite, Gintare Karolina and Roy, Daniel M and Ghahramani, Zoubin},
 title={Training Generative Neural Networks via Maximum Mean Discrepancy Optimization},
 booktitle={UAI},
 year={2015}
}

@article{amfd,
 author={Liu, Wenze and Wang, Xintao and Wan, Pengfei and Yue, Xiangyu},
 title={Amortized Moment Matching for Visual Generation},
 journal={arXiv preprint arXiv:2607.26860},
 year={2026}
}

@article{advfd,
 author={Gao, Mingju and Zhou, Jingkai and Gai, Kun and Yu, Changqian and Tang, Hao},
 title={{AdvFD}: Boosting Visual Generation via Adversarial {Fr{\'e}chet} Distance Loss},
 journal={arXiv preprint arXiv:2608.11205},
 year={2026}
}

@article{zheng2026causal,
  title={{Causal-rCM}: A Unified Teacher-Forcing and Self-Forcing Open Recipe for Autoregressive Diffusion Distillation in Streaming Video Generation and Interactive World Models},
  author={Zheng, Kaiwen and He, Guande and Zhao, Min and Zhang, Jintao and Chen, Huayu and Chen, Jianfei and Lin, Chen-Hsuan and Liu, Ming-Yu and Zhu, Jun and Ma, Qianli},
  journal={arXiv preprint arXiv:2606.25473},
  year={2026}
}

@article{qian2026matrix,
  title={{Matrix-Game 3.5}: Enhancing real-time streaming interactive world models with patch memory},
  author={Qian, Runjia and Wang, Zile and Zhang, Jihai and Zou, Kai and Yu, Wei and Li, Jiaxing and Liu, Zexiang and Li, Yaokun and Kang, Fei and Huang, Kaichen and others},
  journal={arXiv preprint arXiv:2608.29910},
  year={2026}
}

@article{vbenchlong,
  title={{VBench++}: Comprehensive and Versatile Benchmark Suite for Video Generative Models},
  author={Huang, Ziqi and Zhang, Fan and Xu, Xiaojie and He, Yinan and Yu, Jiashuo and Dong, Ziyue and Ma, Qianli and Chanpaisit, Nattapol and Si, Chenyang and Jiang, Yuming and Wang, Yaohui and Chen, Xinyuan and Chen, Ying-Cong and Wang, Limin and Lin, Dahua and Qiao, Yu and Liu, Ziwei},
  journal={IEEE Transactions on Pattern Analysis and Machine Intelligence},
  year={2025},
  doi={10.1109/TPAMI.2025.3633890}
}

@inproceedings{cui2026self,
  title={{Self-Forcing++}: Towards Minute-Scale High-Quality Video Generation},
  author={Cui, Jiaxing and Wu, Jie and Li, Ming and Yang, Tao and Li, Xiaojie and Wang, Rui and Bai, Andrew and Ban, Yuanhao and Hsieh, Cho-Jui},
  booktitle={International Conference on Learning Representations},
  volume={2026},
  pages={85802--85822},
  year={2026}
}

@misc{seedance25,
  author       = {{ByteDance Seed}},
  title        = {One-take Creation, Flexible Referencing: Introducing {Seedance 2.5}},
  year         = {2026},
  month        = jul,
  howpublished = {\url{https://seed.bytedance.com/en/blog/one-take-creation-flexible-referencing-introducing-seedance-2-5}},
  note         = {Accessed: 2026-10-22}
}
\clearpage
\appendix
\raggedbottom
\section{Implementation Details}
\label{sec:implementation}
\subsection{Generator and Optimization}
The generator is initialized from pretrained bidirectional Wan2.1-T2V-1.3B. An 81-frame video is represented by 21 latent frames, partitioned into seven chunks of three latent frames. Attention is bidirectional within each chunk and causal across chunks. The four normalized noise levels are $\{1,0.9375,0.8333,0.625\}$, obtained from the base four-step schedule with timestep shift five. Self-forcing rollouts use the shared exit and gradient truncation in Eq.~\eqref{eq:sf-exit}.

We train on eight NVIDIA B200 GPUs for 250 Adam updates with $(\beta_1,\beta_2)=(0.9,0.95)$ and zero weight decay. The learning rate follows cosine decay from $2.5\times10^{-6}$ to $5\times10^{-7}$ over updates 0--125 and remains at the floor thereafter. The combined generator gradient is clipped to norm 10. Each update uses 128 generated videos for RDM and 32 reference pairs for CCA. All generated samples participate in both MMD estimation and differentiation.

TAEW2.1~\cite{taew,virdm} provides the frozen differentiable decoder for RGB feature losses. Final videos are decoded with the original Wan VAE. Reference features and prompt embeddings are computed offline, and paired CCA targets can be cached.

\subsection{Teacher-Forcing and Self-Forcing Samplers}
CCA uses reference-history predictions with the clean-endpoint map defined in Sec.~\ref{sec:tf}. Independently sampled chunk noise levels and a teacher-forcing mask permit a single masked pass: each target attends to its noisy chunk and preceding clean chunks, excluding its clean target and future clean chunks.

RDM uses self-forcing rollouts generated causally from noise. Each denoising evaluation predicts a clean endpoint; if sampling continues, this prediction is re-noised at the next noise level~\cite{selfforcing,virdm}:
\begin{align}
\hat x^j_{0,s}&=F_\theta\!\left(\sg(x^j_s),\sigma_s;\sg(H^{j-1}),c\right),\\
x^j_{s+1}&=(1-\sigma_{s+1})\sg(\hat x^j_{0,s})
+\sigma_{s+1}\epsilon^j_{s+1}.
\label{eq:rollout}
\end{align}
Re-noising applies for $s<S$. Each chunk returns $\hat x^j=\hat x^j_{0,S}$ and extends the prefix cache for the next chunk. Only the selected evaluations are differentiated. The training distribution $\mathcal P_\theta$ includes prompt sampling, noise, and the shared random exit; evaluation uses all four steps.

\subsection{Feature Definitions}
Layer labels count transformer blocks from one: L$k$ denotes the $k$th block.
\paragraph{VideoMAE v2.}
The VideoMAE v2 encoder has 24 transformer blocks. We discard RGB frame 0 and divide the remaining 80 frames into five chronological clips of 16 frames. Each clip produces 1,568 tokens. CCA averages token-wise cosine distance over all 7,840 tokens. RDM selects 256 ordered positions per clip, yielding 1,280 tokens per video. L16 and L24 refer to the raw post-residual outputs of blocks 16 and 24, respectively. The main visual configuration uses L24 for CCA and L16 for RDM.

\paragraph{V-JEPA 2.1.}
The ablations use the 24-block V-JEPA 2.1 encoder~\cite{vjepa21}. Its input repeats the last frame of the 81-frame video to obtain 82 frames. The representation ablations compare features from blocks 16 and 24.

\paragraph{Latent DiT features.}
The second RDM encoder extracts DiT L9 features from frozen Wan2.1-T2V-14B, using the output of the ninth transformer block (zero-based index 8). It operates on clean normalized video latents and the native Wan text condition. The encoder timestep embedding is fixed to $\tau=300$ for both generated and reference inputs. This $\tau$ is an encoder condition, distinct from the generator's denoising levels. RDM selects 256 positions from the full 32,760-token grid, with 5,120 channels per token. The DiT continuation-alignment ablation uses the full token grid for paired alignment.

For a token sequence of length $P$, fixed indices are selected as $\operatorname{round}(\operatorname{linspace}(0,P-1,K))$, independently within clips when the encoder uses clip segmentation. The selected token matrix is flattened and divided by $\sqrt K$ as in Eq.~\eqref{eq:tokens}. Encoder preprocessing and token layouts remain consistent between generated and reference samples within each branch.

\subsection{Prompt Representations}
\label{sec:prompt-representations}
Our default uses frozen SigLIP2~\cite{siglip2} prompt embeddings with dimension 1,152 and unit-length normalization. Table~\ref{tab:text} replaces these with 2,048-dimensional Qwen3-VL-Embedding-2B~\cite{qwenembedding} embeddings. Both encode prompts only; VideoMAE and the latent DiT supply visual features. The generator's native text-conditioning encoder is unchanged.

For a fixed random subset of reference indices $\mathcal S$ containing up to 2,000 prompts (seed 0), we estimate the text bandwidth as
\begin{equation}
s_c=\left[\operatorname*{median}_{i<j,\ i,j\in\mathcal S}
\|q(c_i)-q(c_j)\|_2^2\right]^{1/2}.
\label{eq:text-bandwidth}
\end{equation}
The median includes all distinct, unordered pairs. The visual bandwidth $s_e$ uses the same rule over reference visual-feature pairs. In implementation, both bandwidths are floored at $10^{-5}$ to avoid division by zero. The text scaling $\beta_e=s_e/s_c$ in Eq.~\eqref{eq:kernel} is the same median-bandwidth ratio used by ViRDM~\cite{virdm}. Changing the prompt encoder requires recomputing $s_c$, the joint landmarks, and the reference kernel-mean coefficients. All bandwidths and reference summaries remain fixed throughout training.

\section{Reference Estimation and Gradient Computation}
\label{sec:estimation}
\subsection{Empirical Objective and Normalization}
\label{sec:empirical-objective}
For a batch of $B$ generated videos, we evaluate all generated--generated kernel interactions, including the diagonal, and approximate reference attraction using a frozen Nystr\"om summary~\cite{rdm,virdm,elastic}. Writing $\hat a_i$ for generated pairs, $\tilde m_e$ for the reference kernel-mean approximation, and $C_e$ for the fixed reference--reference mean gives
\begin{equation}
\widehat{\mathcal D}_e=\frac{1}{B^2}\sum_{i,i'}k_e(\hat a_i,\hat a_{i'})
-\frac{2}{B}\sum_i\tilde m_e(\hat a_i)+C_e.
\label{eq:estimator}
\end{equation}
For the RDM objective, we normalize each encoder's empirical discrepancy before averaging over the encoder set $\mathcal E$:
\begin{equation}
\mathcal L_{\mathrm{RDM}}=\frac{1}{|\mathcal E|}\sum_{e\in\mathcal E}
\frac{\widehat{\mathcal D}_e}{\sg(\widehat{\mathcal D}_e)+\varepsilon_m}.
\label{eq:normalization}
\end{equation}
We use $\varepsilon_m=10^{-4}$.

\subsection{Frozen Reference Kernel Mean}
For joint features $h_e$ in Eq.~\eqref{eq:kernel}, let $h_n^{e,*}=h_e(x_n,c_n)$ be the reference values. Fit $M=4096$ k-means landmarks $U_e$ in this space. With landmark kernel matrix $K_{U_eU_e}$, precompute
\begin{equation}
\begin{aligned}
&b_e=\frac1N\sum_{n=1}^N k_e(U_e,h_n^{e,*}),\\
&\alpha_e=(K_{U_eU_e}+\delta I)^{-1}b_e,\\
&\tilde m_e(h)=k_e(h,U_e)\alpha_e,
\end{aligned}
\label{eq:nystrom}
\end{equation}
where $\delta=10^{-6}$ regularizes the solve and $\alpha_e$ contains unconstrained linear coefficients.

Equation~\eqref{eq:estimator} retains exact generated--generated interactions, including the diagonal, with normalization $B^2$. We also retain the stored reference--reference statistic $C_e$: although its derivative is zero, it contributes to the detached denominator in Eq.~\eqref{eq:normalization}.

We track $\widehat{\mathcal D}_e$ before normalization to assess fitting progress, since the self-normalized value is close to one when $\widehat{\mathcal D}_e\gg\varepsilon_m$.

\subsection{Staged Gradient Computation}
We use the staged VJP strategy of ViRDM~\cite{virdm} to differentiate the frozen feature path in stages and reduce peak graph memory. For an RGB branch with decoder $D$ and encoder $\Phi$, the chain rule is evaluated as
\begin{equation}
\begin{aligned}
g_h&=\nabla_h\mathcal L, & g_v&=J_\Phi^\top g_h,\\
g_x&=J_D^\top g_v, & g_\theta&=J_{G_\theta}^\top g_x.
\end{aligned}
\label{eq:vjp}
\end{equation}
For latent DiT features, the encoder VJP maps directly to video latents. RDM first evaluates cross-video interactions over the complete generated feature population. Feature gradients then propagate through frozen modules and are summed at the latent output before replaying selected-exit generator evaluations. The CCA loss is a sample mean, so its pairwise gradients can be accumulated separately.

\subsection{Joint Update}
A training iteration proceeds as follows:
\begin{enumerate}[leftmargin=*,itemsep=3pt]
\item Sample prompts and noise, select a stochastic denoising exit, and generate the current causal rollout population with detached histories.
\item Extract each RDM representation, evaluate full-batch kernel interactions against the frozen reference summary, and compute feature gradients. Propagate staged VJPs through the encoders, decoder where needed, and replayed generator evaluations to obtain $g_t^{\mathrm{RDM}}$.
\item Sample reference pairs and chunk noise levels. Predict their clean endpoints using the teacher-forcing mask, compute paired token alignment, and obtain $\ell_t$ and $g_t^{\mathrm{CCA}}$.
\item Compute $r_t$ and $\lambda_t$ from the current losses and parameter-gradient norms. Combine the gradients, clip their norm, and perform one optimizer update.
\end{enumerate}
Computing MMD feature gradients over the complete population preserves cross-sample kernel terms during independent sample replay.

\subsection{Gradient Balancing in Practice}
\label{sec:controller-details}
The main-text formulation separates gradient norm matching (GNM) from residual error weighting (REW). In implementation, both are applied through a scalar coefficient on the raw CCA gradient:
\begin{align}
r_t&=r_{\min}+(1-r_{\min})
\clip\!\left(\frac{\ell_t}{\max(\ell_1,\varepsilon_g)},0,1\right),\nonumber\\
\lambda_t&=\sg\!\left[r_t\frac{\|g_t^{\mathrm{RDM}}\|_2}
{\max(\|g_t^{\mathrm{CCA}}\|_2,\varepsilon_g)}\right],\nonumber\\
g_t&=g_t^{\mathrm{RDM}}+\lambda_tg_t^{\mathrm{CCA}},
\label{eq:controller-implementation}
\end{align}
where $r_{\min}=0.1$ and $\varepsilon_g=10^{-12}$ prevents division by zero. Norms cover all trainable generator parameters; $\ell_t$ is the current-batch CCA loss. Controller statistics are treated as constants for the generator update.

For nonzero branch gradients above the numerical floor, $\lambda_tg_t^{\mathrm{CCA}}=r_t\bar g_t^{\mathrm{CCA}}$, so the norm ratio is $r_t$. This relation applies before clipping and Adam preconditioning. The coefficient $\lambda_t$ depends jointly on $r_t$ and the ratio of raw gradient norms.

\section{Additional Ablations}
\label{sec:additional}
Unless otherwise specified, training ablations use 150 updates with RDM and CCA batch sizes of 64 and 32, respectively.
\subsection{Shared Representation Across Components}
\label{sec:shared-representation}
\begin{table}[tbp]
\centering
\footnotesize
\setlength{\tabcolsep}{2.8pt}
\renewcommand{\arraystretch}{1.16}
\caption{Shared representations for continuation alignment and rollout distribution matching. VideoMAE v2 L24 yields the highest Total and Quality, while DiT L9 gives the highest Dynamic.}
\label{tab:shared-space}
\begin{tabularx}{\columnwidth}{>{\raggedright\arraybackslash}Xrrrr}
\toprule
\rowcolor{tablelavender}
\textbf{Shared space} & \textbf{Total} $\uparrow$ & \textbf{Qual.} $\uparrow$ & \textbf{Sem.} $\uparrow$ & \textbf{Dyn.} $\uparrow$ \\
\midrule
VideoMAE v2 L16 & 82.40 & 83.51 & \textbf{77.99} & 54.72 \\
\rowcolor{placeholderfill}
VideoMAE v2 L24 & \textbf{82.69} & \textbf{84.06} & 77.24 & 55.83 \\
V-JEPA 2.1 L16 & 79.69 & 81.06 & 74.21 & 25.28 \\
V-JEPA 2.1 L24 & 80.51 & 81.79 & 75.39 & 25.56 \\
DiT L9 & 82.32 & 83.76 & 76.55 & \textbf{66.94} \\
\bottomrule
\end{tabularx}
\end{table}

Table~\ref{tab:shared-space} sets continuation alignment and rollout distribution matching to the same representation configuration. VideoMAE v2 L24 gives the highest Total and Quality, whereas DiT L9 yields the highest Dynamic Degree. The ranking differs from the separate representation studies, further motivating independent choices for the two components.

\subsection{Training Decoder Configuration}
\label{sec:decoder-ablation}
\begin{table}[tbp]
\centering
\footnotesize
\setlength{\tabcolsep}{2.8pt}
\renewcommand{\arraystretch}{1.16}
\caption{Training decoders with both components using VideoMAE v2 L16. Evaluation uses the original Wan VAE.}
\label{tab:decoder}
\begin{tabularx}{\columnwidth}{>{\raggedright\arraybackslash}Xrrrr}
\toprule
\rowcolor{tablelavender}
\textbf{Training decoder} & \textbf{Total} $\uparrow$ & \textbf{Qual.} $\uparrow$ & \textbf{Sem.} $\uparrow$ & \textbf{Dyn.} $\uparrow$ \\
\midrule
\rowcolor{placeholderfill}
TAEW2.1 & \textbf{82.40} & \textbf{83.51} & 77.99 & \textbf{54.72} \\
Wan VAE (truncated) & 81.70 & 82.62 & \textbf{78.03} & 51.11 \\
\bottomrule
\end{tabularx}
\end{table}

Table~\ref{tab:decoder} compares our TAEW2.1 training configuration, following ViRDM~\cite{virdm}, with the original Wan VAE under temporal-cache gradient truncation. TAEW retains the full temporal decoder VJP, while staged VJPs manage differentiation through the frozen feature path. Replacing this configuration with the truncated Wan VAE reduces Total from 82.40 to 81.70 and Dynamic from 54.72 to 51.11. This comparison changes both the decoder and its temporal gradient path, so the scores characterize the complete training configuration. Evaluation uses the original Wan VAE in both cases.

\subsection{Reducing the Denoising Budget}
\label{sec:fewer-steps}
We evaluate one- and two-step variants using our main representation configuration: \vm L24 for CCA and \vm L16 plus DiT L9 for RDM. All variants use 150 updates and the same 944 expanded prompts, with five videos per prompt. RDM rollouts and inference retain four denoising steps for the first chunk and use the specified budget for later chunks. The one- and two-step noise grids are $\{1\}$ and $\{1,0.8333\}$, respectively, after timestep shift 5.

CCA independently samples each chunk's noise level from the four-step grid $\{1,0.9375,0.8333,0.625\}$, regardless of the rollout budget. All clean endpoints are predicted in one masked forward pass (Sec.~\ref{sec:tf}).

\begin{table}[htbp]
\centering
\footnotesize
\setlength{\tabcolsep}{1.7pt}
\renewcommand{\arraystretch}{1.16}
\caption{Denoising-step ablation of \method after 150 updates. Steps refer to later chunks in RDM rollouts and inference; the first chunk uses four steps. CCA retains the four-level noise grid for all chunks.}
\label{tab:fewstep}
\begin{tabularx}{\columnwidth}{@{}>{\raggedright\arraybackslash}p{.26\columnwidth}*{4}{>{\raggedleft\arraybackslash}X}@{}}
\toprule
\rowcolor{tablelavender}
\textbf{Steps} & \textbf{Total} $\uparrow$ & \textbf{Quality} $\uparrow$ & \textbf{Semantic} $\uparrow$ & \textbf{Dynamic} $\uparrow$ \\
\midrule
\rowcolor{black!4}
4 & \textbf{84.11} & \textbf{85.47} & \textbf{78.67} & \textbf{74.72} \\
2 & 83.77 & 85.20 & 78.03 & 70.00 \\
1 & 81.58 & 82.75 & 76.94 & 69.17 \\
\bottomrule
\end{tabularx}
\end{table}

Table~\ref{tab:fewstep} shows that the two-step variant achieves 83.77 Total and 85.20 Quality, close to the four-step model's 84.11 and 85.47, with 78.03 Semantic and 70.00 Dynamic. The one-step variant achieves 81.58 Total, 82.75 Quality, 76.94 Semantic, and 69.17 Dynamic. Thus, one-step rollouts retain substantial motion but have lower Quality and Semantic than the two- and four-step variants.

\subsection{Optical-Flow Regularization}
\label{sec:flow-ablation}
We add ViRDM's optical-flow magnitude regularizer~\cite{virdm} to the common ablation configuration, using \vm L16 for both components and four denoising steps per chunk. We keep the 150-update training protocol and the 4,720-video evaluation. The generated video is decoded with TAEW2.1 and subsampled to 8 frames/s, yielding 40 adjacent frame pairs. A frozen DenseFlowModel estimates optical flow with 20 iterations. For each pair, we average the largest 5\% of pixel-wise flow magnitudes, then apply a hinge to the tenth-largest pair score:
\begin{align}
s_t(v)&=\operatorname{MeanTop}_{5\%}
  \bigl(\{\|F_t(p)\|_2\}_p\bigr),\nonumber\\
\mathcal L_{\mathrm{flow}}&=
  \mathbb E_v\bigl[\,11.25-S_{10}(v)\,\bigr]_+,
\label{eq:flow-ablation}
\end{align}
where $F_t(p)$ is the flow at pixel $p$, $S_{10}$ is the tenth-largest value among the 40 pair scores, and the threshold is measured in pixels. We add $w\mathcal L_{\mathrm{flow}}$ to the RDM objective for $w\in\{10^{-4},5\times10^{-4}\}$. Gradient norm matching then uses the combined rollout gradient $\nabla_\theta(\mathcal L_{\mathrm{RDM}}+w\mathcal L_{\mathrm{flow}})$ as its reference when scaling the continuation gradient.

\begin{table}[htbp]
\centering
\footnotesize
\setlength{\tabcolsep}{2.8pt}
\renewcommand{\arraystretch}{1.16}
\caption{Optical-flow regularization with both components using VideoMAE v2 L16. Increasing the weight improves Total and Dynamic.}
\label{tab:flow-scores}
\begin{tabularx}{\columnwidth}{>{\raggedright\arraybackslash}Xrrrr}
\toprule
\rowcolor{tablelavender}
\textbf{Flow weight $w$} & \textbf{Total} $\uparrow$ & \textbf{Qual.} $\uparrow$ & \textbf{Sem.} $\uparrow$ & \textbf{Dyn.} $\uparrow$ \\
\midrule
0 & 82.40 & 83.51 & 77.99 & 54.72 \\
$10^{-4}$ & 82.93 & 84.12 & \textbf{78.19} & 77.50 \\
$5\times10^{-4}$ & \textbf{83.46} & \textbf{84.87} & 77.84 & \textbf{90.56} \\
\bottomrule
\end{tabularx}
\end{table}

\begin{table}[htbp]
\centering
\footnotesize
\setlength{\tabcolsep}{2.8pt}
\renewcommand{\arraystretch}{1.16}
\caption{Full VBench dimension scores for optical-flow regularization. Columns give flow weight $w$. Scores are multiplied by 100; higher is better for every metric.}
\label{tab:flow-dimensions}
\begin{tabularx}{\columnwidth}{>{\raggedright\arraybackslash}Xrrr}
\toprule
\rowcolor{tablelavender}
\textbf{Metric} & 0 & $10^{-4}$ & $5\times10^{-4}$ \\
\midrule
Subject Consistency & \textbf{95.69} & 95.51 & 95.47 \\
Background Consistency & \textbf{95.62} & 94.92 & 94.87 \\
Temporal Flickering & \textbf{99.15} & 97.72 & 97.73 \\
Motion Smoothness & \textbf{98.77} & 98.44 & 98.32 \\
Dynamic Degree & 54.72 & 77.50 & \textbf{90.56} \\
Aesthetic Quality & \textbf{65.07} & 64.85 & 64.72 \\
Imaging Quality & \textbf{66.99} & 65.92 & 64.91 \\
Object Class & \textbf{96.14} & 95.95 & 95.25 \\
Multiple Objects & \textbf{81.62} & 79.73 & 79.16 \\
Human Action & 94.00 & \textbf{94.60} & 94.40 \\
Color & 79.11 & \textbf{81.02} & 80.83 \\
Spatial Relationship & 75.63 & \textbf{76.82} & 76.44 \\
Scene & 52.99 & \textbf{54.03} & 53.44 \\
Temporal Style & \textbf{23.76} & 23.72 & 23.75 \\
Appearance Style & \textbf{21.19} & 21.01 & 20.89 \\
Overall Consistency & \textbf{26.04} & 25.91 & 25.89 \\
\bottomrule
\end{tabularx}
\end{table}

In our tested configuration, optical-flow regularization improves Total and Dynamic Degree (Table~\ref{tab:flow-scores}) but reduces Temporal Flickering, Motion Smoothness, Aesthetic Quality, and Imaging Quality scores (Table~\ref{tab:flow-dimensions}). Visual inspection also reveals implausible shaking. These observations suggest that rewarding displacement magnitude can increase measured motion without ensuring plausible temporal dynamics. We therefore omit this auxiliary objective from our main configuration.

\subsection{Token Representations versus Global Pooling}
\label{sec:pooling-ablation}
Table~\ref{tab:pooling} compares two representations for rollout distribution matching, using \vm L16 for both training components and retaining full-token cosine supervision for continuation alignment. Global pooling averages all 7,840 video tokens into a single 1,024-dimensional vector. The token representation retains 256 fixed, ordered tokens from each of five clips, then flattens the resulting 1,280 tokens and divides by $\sqrt{1280}$. Reference summaries and kernel bandwidths are recomputed for each representation. Both variants use 150 joint updates and the same 4,720-video evaluation protocol.

Retaining tokens improves Total, Quality, and Semantic by 1.65, 1.67, and 1.60 points, respectively, with similar Dynamic Degree (54.72 vs.\ 55.56). These gains support preserving token-level information for distribution matching, primarily benefiting overall quality and semantic alignment.

\begin{table}[htbp]
\centering
\footnotesize
\setlength{\tabcolsep}{1.7pt}
\renewcommand{\arraystretch}{1.16}
\caption{Pooling ablation for rollout distribution matching, with VideoMAE v2 L16 for both training components.}
\label{tab:pooling}
\begin{tabularx}{\columnwidth}{@{}>{\raggedright\arraybackslash}p{.26\columnwidth}*{4}{>{\raggedleft\arraybackslash}X}@{}}
\toprule
\rowcolor{tablelavender}
\textbf{Representation} & \textbf{Total} $\uparrow$ & \textbf{Quality} $\uparrow$ & \textbf{Semantic} $\uparrow$ & \textbf{Dynamic} $\uparrow$ \\
\midrule
Global pooling & 80.75 & 81.84 & 76.39 & \textbf{55.56} \\
\rowcolor{black!4}
Tokens & \textbf{82.40} & \textbf{83.51} & \textbf{77.99} & 54.72 \\
\bottomrule
\end{tabularx}
\end{table}

\subsection{V-JEPA Continuation Features}
\label{sec:vjepa-continuation}
Table~\ref{tab:appendix-vjepa-continuation} compares V-JEPA 2.1 L16 and L24 for continuation alignment while fixing rollout distribution matching to \vm L24. L24 improves all four metrics over L16, raising Total from 81.14 to 81.62 and Dynamic from 28.61 to 35.56. This agrees with the preference for final-layer continuation features observed when distribution matching uses \vm L16 (Table~\ref{tab:tf-space}).

\begin{table}[H]
\centering
\footnotesize
\setlength{\tabcolsep}{2.8pt}
\renewcommand{\arraystretch}{1.16}
\caption{V-JEPA 2.1 features for continuation alignment, with rollout distribution matching fixed to VideoMAE v2 L24.}
\label{tab:appendix-vjepa-continuation}
\begin{tabularx}{\columnwidth}{>{\raggedright\arraybackslash}Xrrrr}
\toprule
\rowcolor{tablelavender}
\textbf{CCA representation} & \textbf{Total} $\uparrow$ & \textbf{Qual.} $\uparrow$ & \textbf{Sem.} $\uparrow$ & \textbf{Dyn.} $\uparrow$ \\
\midrule
V-JEPA 2.1 L16 & 81.14 & 82.01 & 77.65 & 28.61 \\
\rowcolor{black!4}
V-JEPA 2.1 L24 & \textbf{81.62} & \textbf{82.47} & \textbf{78.24} & \textbf{35.56} \\
\bottomrule
\end{tabularx}
\end{table}

\subsection{Qualitative Effects of CCA and RDM}
\label{sec:component-ablation}
\begin{figure*}[t]
\centering
\includegraphics[width=0.95\textwidth]{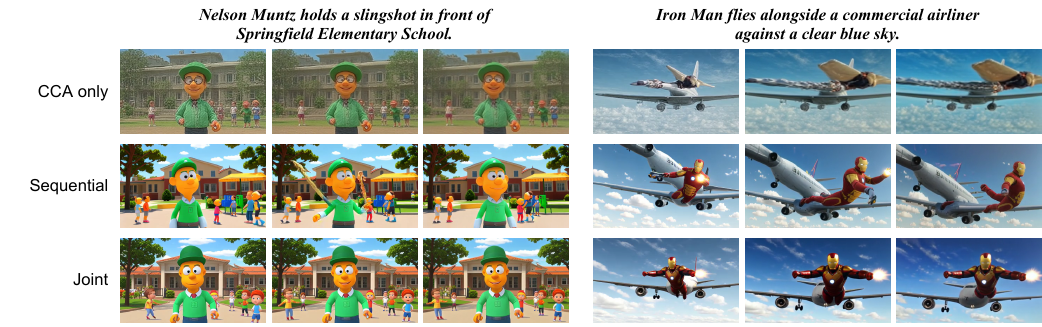}
\caption{\textbf{Qualitative effects of the training schedule.} Rows show CCA-only, sequential CCA initialization followed by RDM-only, and joint training. CCA-only and joint training use 150 updates; sequential training uses 150+150. All rows use frames 0/16/32 for the school case and 0/40/80 for the airliner case. Prompts are abbreviated.}
\label{fig:component-training}
\end{figure*}

Figure~\ref{fig:component-training} compares three configurations from Table~\ref{tab:component-ablation}. CCA-only samples retain content across frames but show blurred detail and distorted structure. Sequential training improves appearance, yet exhibits transient artifacts in the school example and body deformation in the flying-character example. Joint training preserves cleaner structures across the displayed frames.

\subsection{CCA Batch Size Sensitivity}
\label{sec:cca-batch-size}
Table~\ref{tab:cca-batch-size} varies the CCA batch size while using \vm L16 for both components. Among batch sizes 16, 32, and 64, 32 achieves the highest Total, Quality, and Dynamic Degree, whereas 16 gives the highest Semantic score. We therefore retain 32 as the best overall setting among the tested configurations.

\begin{table}[htbp]
\centering
\footnotesize
\setlength{\tabcolsep}{1.7pt}
\renewcommand{\arraystretch}{1.16}
\caption{Sensitivity to the CCA batch size, using VideoMAE v2 L16 for both CCA and RDM. The default batch size is 32.}
\label{tab:cca-batch-size}
\begin{tabularx}{\columnwidth}{@{}>{\raggedright\arraybackslash}p{.26\columnwidth}*{4}{>{\raggedleft\arraybackslash}X}@{}}
\toprule
\rowcolor{tablelavender}
\textbf{CCA batch size} & \textbf{Total} $\uparrow$ & \textbf{Quality} $\uparrow$ & \textbf{Semantic} $\uparrow$ & \textbf{Dynamic} $\uparrow$ \\
\midrule
16 & 81.84 & 82.64 & \textbf{78.61} & 38.06 \\
\rowcolor{black!4}
32 & \textbf{82.40} & \textbf{83.51} & 77.99 & \textbf{54.72} \\
64 & 82.03 & 83.21 & 77.32 & 51.67 \\
\bottomrule
\end{tabularx}
\end{table}

\subsection{Additional Representation Combinations}
\label{sec:additional-combinations}
Table~\ref{tab:additional-combinations} extends the main comparison (Table~\ref{tab:representation-combinations}) to two CCA feature sets and four RDM feature sets. All use 150 updates, SigLIP2 prompt embeddings, GNM and REW, and RDM/CCA batch sizes of 64/32.

\begin{table}[htbp]
\centering
\footnotesize
\setlength{\tabcolsep}{1.5pt}
\renewcommand{\arraystretch}{1.16}
\caption{Additional representation combinations after 150 updates. $V_k$: VideoMAE v2 L$k$; $D_9$: DiT L9. Gray marks the representation configuration used by our main model.}
\label{tab:additional-combinations}
\begin{tabularx}{\columnwidth}{*{2}{>{\centering\arraybackslash}X}rrrr}
\toprule
\rowcolor{tablelavender}
\textbf{Continuation} & \textbf{Distribution} & \textbf{Total} $\uparrow$ & \textbf{Qual.} $\uparrow$ & \textbf{Sem.} $\uparrow$ & \textbf{Dyn.} $\uparrow$ \\
\midrule
$V_{24}$ & $V_{16}$ & 82.80 & 84.00 & 78.01 & 58.06 \\
$V_{24}$ & $D_{9}$ & 83.49 & 84.96 & 77.58 & 68.61 \\
\rowcolor{black!4}
$V_{24}$ & $V_{16}+D_{9}$ & \textbf{84.11} & \textbf{85.47} & 78.67 & 74.72 \\
$V_{24}$ & $V_{16}+D_{9}+V_{24}$ & 83.86 & 85.31 & 78.05 & \textbf{77.78} \\
\addlinespace[3pt]
$V_{24}+D_{9}$ & $V_{16}$ & 83.35 & 84.36 & \textbf{79.32} & 56.39 \\
$V_{24}+D_{9}$ & $D_{9}$ & 83.04 & 84.37 & 77.72 & 65.00 \\
$V_{24}+D_{9}$ & $V_{16}+D_{9}$ & 83.68 & 84.99 & 78.43 & 71.94 \\
$V_{24}+D_{9}$ & $V_{16}+D_{9}+V_{24}$ & 83.46 & 84.71 & 78.45 & 75.00 \\
\bottomrule
\end{tabularx}
\end{table}

With RDM fixed to $V_{16}$, adding $D_9$ to $V_{24}$ for CCA raises Total from 82.80 to 83.35 and improves Quality and Semantic, but slightly reduces Dynamic. For $D_9$-only RDM, the addition instead lowers Total (83.49 to 83.04), Quality, and Dynamic, with a small Semantic gain; for $V_{16}+D_9$ RDM, all four metrics decline. Thus, the benefit of combining CCA representations depends on the RDM feature set.

With CCA fixed to $V_{24}$, adding $V_{24}$ to the $V_{16}+D_9$ RDM feature set increases Dynamic from 74.72 to 77.78 but reduces Total, Quality, and Semantic. The selected $V_{24}$ / $V_{16}+D_9$ configuration retains the highest Total and Quality in this extended comparison, while other combinations favor Semantic or Dynamic.

\section{DiT Feature Studies}
\label{sec:dit-feature-studies}
\subsection{Clean versus Noised DiT Features}
\label{sec:clean-noised-features}
We examine how input noising affects frozen DiT L9 features from Wan2.1-T2V-14B, keeping the encoder time condition fixed at $\tau=300$. We compare clean latents $x$ with noised latents $0.7x+0.3\epsilon$, measuring the mean token-wise cosine distance between original and perturbed videos over all 32,760 tokens. Cross-video splicing uses 32 disjoint video pairs, replacing the final 41 frames of one video with those of another. Temporal freezing uses 64 videos, uniformly retaining 50\% of frames and repeating them to restore 81 frames. Original and perturbed inputs share the prompt and, when noised, the same noise realization. We average the noised results over three noise seeds.

Table~\ref{tab:clean-noised-features} shows that forward noising reduces the feature response by 27.3\% for cross-video splicing and by 22.7\% for temporal freezing. Clean inputs produce stronger responses in both diagnostics, indicating greater sensitivity to the tested content and temporal changes. These results support our use of clean inputs for DiT feature extraction.

\begin{table}[htbp]
\centering
\footnotesize
\setlength{\tabcolsep}{2.4pt}
\renewcommand{\arraystretch}{1.16}
\caption{Response of clean and noised DiT L9 features at $\tau=300$. Distances are multiplied by 100; Reduction gives the relative decrease after noising.}
\label{tab:clean-noised-features}
\begin{tabularx}{\columnwidth}{@{}>{\raggedright\arraybackslash}Xrrr@{}}
\toprule
\rowcolor{tablelavender}
\textbf{Perturbation} & \textbf{Clean} & \textbf{Noised} & \textbf{Reduction} \\
\midrule
Cross-video splicing & 24.75 & 17.99 & 27.29\% \\
Temporal freezing: 50\% retained & 4.31 & 3.33 & 22.68\% \\
\bottomrule
\end{tabularx}
\end{table}

\subsection{DiT Layer and Timestep Selection}
\label{sec:dit-layer-selection}
Table~\ref{tab:dit-layer-selection} compares DiT layers and encoder time conditions $\tau$, using clean latents and the 150-update protocol in Sec.~\ref{sec:ablations}, with the other component fixed to \vm L16. At $\tau=300$, L25 achieves the highest Total, Quality, and Semantic for CCA, while L9 retains higher Dynamic. For RDM, L9 outperforms L25 and L40 on all four metrics. The strongest results come from intermediate features for CCA and shallow features for RDM, supporting the use of shallow-to-intermediate DiT representations over the deepest layer.

With L9 fixed, $\tau=300$ achieves the highest Total and Quality in both components. The overall results favor encoder time conditions associated with low to moderate noise levels for both tasks. At this layer, CCA attains its highest Semantic at $\tau=0$ and Dynamic at $\tau=900$; for RDM, $\tau=0$ gives slightly higher Dynamic than $\tau=300$. We retain L9 with $\tau=300$ in our main configuration for its strong performance and lower feature-extraction memory requirements than deeper DiT layers.

\begin{table}[t]
\centering
\footnotesize
\setlength{\tabcolsep}{2.2pt}
\renewcommand{\arraystretch}{1.16}
\caption{DiT layer and encoder time condition $\tau$ for CCA and RDM. All DiT inputs are clean latents; the other component uses VideoMAE v2 L16. Gray marks L9 at $\tau=300$; best scores are bold within each component.}
\label{tab:dit-layer-selection}
\begin{tabularx}{\columnwidth}{@{}*{2}{>{\centering\arraybackslash}X}rrrr@{}}
\toprule
\rowcolor{tablelavender}
\textbf{Layer} & $\boldsymbol\tau$ & \textbf{Total} $\uparrow$ & \textbf{Quality} $\uparrow$ & \textbf{Semantic} $\uparrow$ & \textbf{Dynamic} $\uparrow$ \\
\midrule
\multicolumn{6}{@{}l}{\emph{CCA: RDM uses VideoMAE v2 L16}} \\
L9 & 0 & 81.88 & 82.54 & 79.25 & 44.72 \\
\rowcolor{black!4}
L9 & 300 & 82.56 & 83.53 & 78.68 & 58.89 \\
L9 & 600 & 82.10 & 82.88 & 78.99 & 48.33 \\
L9 & 900 & 82.02 & 83.02 & 78.05 & \textbf{65.28} \\
L25 & 300 & \textbf{82.90} & \textbf{83.78} & \textbf{79.36} & 56.39 \\
L40 & 300 & 82.15 & 82.93 & 79.04 & 46.39 \\
\addlinespace[3pt]
\multicolumn{6}{@{}l}{\emph{RDM: CCA uses VideoMAE v2 L16}} \\
L9 & 0 & 81.98 & 83.20 & 77.10 & \textbf{52.78} \\
\rowcolor{black!4}
L9 & 300 & \textbf{82.45} & \textbf{83.53} & \textbf{78.13} & 52.22 \\
L9 & 600 & 81.65 & 82.71 & 77.41 & 41.67 \\
L9 & 900 & 78.92 & 79.72 & 75.68 & 25.28 \\
L25 & 300 & 79.21 & 80.38 & 74.55 & 15.56 \\
L40 & 300 & 80.48 & 82.63 & 71.90 & 43.89 \\
\bottomrule
\end{tabularx}
\end{table}

\section{Additional Main-Model Results}
\label{sec:additional-main-results}
\setlength{\textfloatsep}{10pt plus 2pt minus 2pt}
\setlength{\floatsep}{6pt plus 2pt minus 2pt}
\setlength{\intextsep}{8pt plus 2pt minus 2pt}
\subsection{Dynamic Degree in the Main Comparison}
\label{sec:main-dynamic}
Table~\ref{tab:main-dynamic} reports Dynamic Degree for the main comparison. We evaluate \method and Elastic Forcing's released three-encoder checkpoint on the 944 expanded VBench prompts released by Causal Forcing, generating five videos per prompt. ViRDM uses the reported score of its variant without optical-flow regularization.

\begin{table}[H]
\centering
\footnotesize
\setlength{\tabcolsep}{2.8pt}
\renewcommand{\arraystretch}{1.16}
\caption{Dynamic Degree in the main comparison.}
\label{tab:main-dynamic}
\begin{tabularx}{\columnwidth}{@{}l*{3}{>{\centering\arraybackslash}X}@{}}
\toprule
\rowcolor{tablelavender}
 & \textbf{ViRDM} & \textbf{Elastic Forcing} & \textbf{\methodname} \\
\midrule
Dynamic $\uparrow$ & 48.61 & 63.33 & \textbf{75.56} \\
\bottomrule
\end{tabularx}
\end{table}

\subsection{RDM Batch Size in the Main Model}
\label{sec:main-rdm-batch-size}
For each RDM batch size, we train for 250 updates and report all metrics from the checkpoint with the highest VBench Total within this budget. Batch size 128 achieves the highest scores on all four metrics (Table~\ref{tab:main-rdm-batch-size}), and we adopt it for our main configuration.

\begin{table}[t]
\centering
\footnotesize
\setlength{\tabcolsep}{2pt}
\renewcommand{\arraystretch}{1.16}
\caption{RDM batch size with VideoMAE v2 L24 for CCA and VideoMAE v2 L16 plus DiT L9 for RDM. Each row reports the highest-Total checkpoint within the first 250 updates; CCA batch size is 32. Gray marks the main setting.}
\label{tab:main-rdm-batch-size}
\begin{tabularx}{\columnwidth}{@{}cc*{4}{>{\raggedleft\arraybackslash}X}@{}}
\toprule
\rowcolor{tablelavender}
\textbf{Batch size} & \textbf{Update} & \textbf{Total} $\uparrow$ & \textbf{Quality} $\uparrow$ & \textbf{Semantic} $\uparrow$ & \textbf{Dynamic} $\uparrow$ \\
\midrule
64 & 150 & 84.11 & 85.47 & 78.67 & 74.72 \\
\rowcolor{black!4}
128 & 250 & \textbf{84.63} & \textbf{86.01} & \textbf{79.08} & \textbf{75.56} \\
256 & 175 & 83.87 & 85.24 & 78.36 & 71.94 \\
512 & 75 & 83.63 & 85.00 & 78.15 & 65.28 \\
\bottomrule
\end{tabularx}
\end{table}

\subsection{Temporal Continuity}
\label{sec:temporal-continuity}
We compare our single-stage model with the released ViRDM~\cite{virdm} model, which uses causal ODE initialization. Both exclude optical-flow regularization. Boundary diagnostics use 64 fixed paired samples and the generation settings in Sec.~\ref{sec:experimental-setup}.

\begin{table}[ht]
\centering
\footnotesize
\setlength{\tabcolsep}{3pt}
\renewcommand{\arraystretch}{1.16}
\caption{Temporal continuity of the ODE-initialized ViRDM pipeline and our single-stage pipeline, both without optical-flow regularization. Brackets denote 95\% prompt-bootstrap confidence intervals. Boundary ratios closer to one are better.}
\label{tab:temporal-continuity}
\begin{tabularx}{\columnwidth}{@{}l>{\centering\arraybackslash}Xc@{}}
\toprule
\rowcolor{tablelavender}
\textbf{Method} & \shortstack{\textbf{Chunk boundary}\\\textbf{ratio} ($\approx 1$)} & \shortstack{\textbf{Motion}\\\textbf{Smoothness} $\uparrow$} \\
\midrule
ViRDM~\cite{virdm} & 1.393 [1.332, 1.461] & 98.12 \\
\rowcolor{black!4}
\methodname & \textbf{1.040 [0.973, 1.129]} & \textbf{98.52} \\
\bottomrule
\end{tabularx}
\end{table}

\paragraph{Chunk boundary ratio.}
On float RGB before video compression, we compute the mean absolute difference between consecutive frames over all pixels and color channels. Boundary transitions enter frames $\mathcal B=\{9,21,33,45,57,69\}$ (zero-based); the remaining transitions form $\mathcal I$. For model $m$, let $\bar d_{\mathcal B}^{(m)}$ and $\bar d_{\mathcal I}^{(m)}$ denote these differences averaged within each set and equally across samples. Applying the same partition to reference videos, we define
\begin{equation}
J^{(m)}=\frac{\bar d_{\mathcal B}^{(m)}/\bar d_{\mathcal I}^{(m)}}
{\bar d_{\mathcal B}^{(\mathrm{GT})}/\bar d_{\mathcal I}^{(\mathrm{GT})}},
\label{eq:chunk-boundary-ratio}
\end{equation}
where GT denotes reference videos. Values closer to one indicate a boundary-to-interior variation pattern closer to the references.

\paragraph{Results.}
\method achieves a boundary ratio closer to one than ViRDM (1.040 vs.\ 1.393; Table~\ref{tab:temporal-continuity}). VBench Motion Smoothness~\cite{vbench}, evaluated under the expanded-prompt protocol in Sec.~\ref{sec:experimental-setup}, is also higher (98.52 vs.\ 98.12). These results support smooth chunk transitions without a separate causal initialization stage.

\subsection{Sources of Baseline Scores and Timings}
\label{sec:baseline-score-sources}
For the Total, Quality, and Semantic scores in Table~\ref{tab:main}, we use the values reported by Causal Forcing~\cite{causalforcing} for LTX-Video-1.9B, Wan2.1-1.3B, NOVA, Pyramid Flow, SkyReels-V2-1.3B, MAGI-1-4.5B, CausVid, Self Forcing, and Causal Forcing itself. Scores for LongLive and Causal-rCM follow Causal-rCM~\cite{zheng2026causal}. ViRDM and Elastic Forcing scores follow their respective reports~\cite{virdm,elastic}. Model citations in Table~\ref{tab:main} identify the original methods.

\paragraph{Inference timing.}
For throughput and latency in Table~\ref{tab:main}, we use the values reported by Causal Forcing~\cite{causalforcing} for LTX-Video-1.9B, Wan2.1-1.3B, NOVA, Pyramid Flow, SkyReels-V2-1.3B, and MAGI-1-4.5B. The four-step causal models use the single-B200 measurements described in Sec.~\ref{sec:experimental-setup}.

\section{Long-Video Extension}
\label{sec:long-video-details}
\subsection{Training with Short Reference Videos}
\label{sec:long-video-training}
Starting from the 150-update short-video checkpoint (Table~\ref{tab:representation-combinations}), we perform 100 additional updates using the same 6,505 short reference videos.

\paragraph{Windows on long generated histories.}
We uniformly sample 21-latent-frame windows at chunk-aligned positions within a 240-latent-frame rollout, corresponding to approximately five-second windows over a 60-second horizon. Prefix generation is detached; window sampling and differentiation follow Eq.~\eqref{eq:sf-exit}. Training and inference use a bounded context of three sink latent frames, six recent history frames, and the current three-frame chunk, with FIFO eviction and top-aligned RoPE.

\paragraph{Pooled distribution matching.}
Long generated histories lack paired reference continuations.

We match the distribution of their sampled windows to the short-reference distribution using globally pooled VideoMAE v2 L16 and DiT L9 features. For each encoder $e$, we average its full token grid:
\begin{equation}
\begin{aligned}
z_e^{\mathrm{pool}}(w,c)&=\frac{1}{P_e}\sum_{p=1}^{P_e}\Phi_e(w,c)_p,\\
\mathcal D_e^{\mathrm{long}}(\theta)&=\operatorname{MMD}_{k_e}^{2}
\bigl(\mathcal P_\theta^{\mathrm{win}},\mathcal Q_{\mathcal D}\bigr),
\end{aligned}
\label{eq:long-pooled-rdm}
\end{equation}
where $\mathcal P_\theta^{\mathrm{win}}$ includes random window positions and $\mathcal Q_{\mathcal D}$ is the short-reference distribution. Pooling follows temporal encoding and removes fixed token coordinates from the readout. We use the SigLIP2 joint kernel in Eq.~\eqref{eq:kernel}, recompute bandwidths and reference summaries from pooled features, and retain the discrepancy normalization in Eq.~\eqref{eq:normalization}.

CCA continues full-token VideoMAE v2 L24 alignment on paired short references, with the same GNM/REW update rule. Adaptation uses RDM/CCA batch sizes of 64/32 and a constant learning rate of $5\times10^{-7}$; Adam state and the initial REW residual are reset at the start. Remaining implementation details follow Appendices~\ref{sec:implementation}--\ref{sec:estimation}.

\subsection{Evaluation Protocol}
\label{sec:long-video-evaluation}
We generate one 50-second video per prompt with a fixed seed, using the first 128 expanded MovieGenVideoBench prompts released by Self Forcing~\cite{selfforcing}. VBench-Long~\cite{vbenchlong} runs in custom-input mode with scene-cut splitting disabled and its official consistency aggregation. Table~\ref{tab:long-video} compares six metrics with the reported Self-Forcing++~\cite{cui2026self} results on MovieGen-128 at the same duration.

\section{Training on Closed-Source Model Outputs}
\label{sec:closed-source-data}
\method learns from reference videos without requiring access to the reference generator's weights or online diffusion-score evaluations. This enables training on outputs from closed-source video generators and allows the reference corpus to be curated before training.

We construct a corpus of 1,171 Seedance 2.5-generated videos~\cite{seedance25} with substantial motion and English descriptions, manually excluding videos with implausible motion. For each 5--20-second source clip, we sample frames across its full duration to construct a five-second reference sequence. Both CCA and RDM are trained on this corpus for 150 updates, using \vm L16 for both components and RDM/CCA batch sizes of 64/32. RDM uses a reference bank containing the complete curated corpus. We compare this model with the corresponding L16/L16 model trained on the original 6,505 reference videos, evaluating both with the full VBench suite.

\begin{table}[t]
\centering
\small
\setlength{\tabcolsep}{3pt}
\renewcommand{\arraystretch}{1.12}
\caption{Full VBench scores of \method trained on the indicated reference data. Scores are percentages.}
\label{tab:closed-source-data}
\begin{tabularx}{\columnwidth}{@{}l>{\centering\arraybackslash}X>{\centering\arraybackslash}X@{}}
\toprule
\rowcolor{tablelavender}
\textbf{Metric} & \textbf{Original data} & \textbf{Seedance 2.5} \\
\midrule
Dynamic Degree $\uparrow$ & 54.72 & \textbf{62.50} \\
Motion Smoothness $\uparrow$ & 98.77 & \textbf{98.90} \\
Subject Consistency $\uparrow$ & \textbf{95.69} & 93.81 \\
Background Consistency $\uparrow$ & \textbf{95.62} & 93.80 \\
\bottomrule
\end{tabularx}
\end{table}

Table~\ref{tab:closed-source-data} shows that Dynamic Degree increases by 7.78 points, while Motion Smoothness improves slightly by 0.13. Subject and Background Consistency decrease by 1.88 and 1.82 points, respectively. These results demonstrate that \method can learn from curated closed-source video outputs, providing a path to benefit from more capable generators and larger curated reference corpora.

\end{document}